\documentclass[journal,12pt,onecolumn,draftclsnofoot]{IEEEtran}

\usepackage{amsmath,amsfonts}
\usepackage{bm}
\usepackage{xcolor}
\usepackage{booktabs}
\usepackage{array}
\usepackage{caption}
\usepackage{algorithm}
\usepackage{algorithmicx}
\usepackage{algpseudocode}
\usepackage{array}
\usepackage[caption=false,font=footnotesize,labelfont=sf,textfont=sf]{subfig}
\usepackage{textcomp}
\usepackage{stfloats}
\usepackage{url}
\usepackage{graphicx}
\usepackage{cite}
\usepackage{color}
\usepackage{hyperref}
\usepackage{multirow}
\hypersetup{pdfborder={0 0 0}}
\usepackage{overpic}

\ifCLASSINFOpdf
\else
\fi
\begin{document}
%
\title{AI Flow: Perspectives, Scenarios, and Approaches}
%
%
%

\author{Hongjun~An, Wenhan~Hu, Sida~Huang, Siqi~Huang, Ruanjun~Li, Yuanzhi~Liang, Jiawei~Shao, Yiliang~Song, Zihan~Wang, Cheng~Yuan, Chi~Zhang, Hongyuan~Zhang, Wenhao~Zhuang, Xuelong~Li$^{*}$ \\
\IEEEauthorblockA{Institute of Artificial Intelligence (TeleAI), China Telecom}
\thanks{
The authors are affiliated with the Institute of Artificial Intelligence (TeleAI), China Telecom, China. 
Author names are listed alphabetically by surname.
This work was conducted at TeleAI, facilitated by Dr. Jiawei Shao (e-mail: shaojw2@chinatelecom.cn) under the leadership of Prof. Xuelong Li.
The corresponding author is Prof. Xuelong Li (e-mail: xuelong\_li@ieee.org), the CTO and Chief Scientist of China Telecom.
}
}
\maketitle

\vspace{-1.5cm}

\begin{abstract}

Pioneered by the foundational information theory by Claude Shannon and the visionary framework of machine intelligence by Alan Turing, the convergent evolution of information and communication technologies (IT/CT) has created an unbroken wave of connectivity and computation. 
This synergy has sparked a technological revolution, now reaching its peak with large artificial intelligence (AI) models that are reshaping industries and redefining human-machine collaboration.
However, the realization of ubiquitous intelligence faces considerable challenges due to substantial resource consumption in large models and high communication bandwidth demands.
To address these challenges, \emph{AI Flow} has been introduced as a multidisciplinary framework that integrates cutting-edge IT and CT advancements, with a particular emphasis on the following three key points.
\textbf{First, \emph{device-edge-cloud framework} serves as the foundation}, which integrates end devices, edge servers, and cloud clusters to optimize scalability and efficiency for low-latency model inference.
\textbf{Second, we introduce the concept of \emph{familial models}, which refers to a series of different-sized models with aligned hidden features}, enabling effective collaboration and the flexibility to adapt to varying resource constraints and dynamic scenarios.
\textbf{Third, \emph{connectivity- and interaction-based intelligence emergence} is a novel paradigm of AI Flow. }
By leveraging communication networks to enhance connectivity, the collaboration among AI models across heterogeneous nodes achieves emergent intelligence that surpasses the capability of any single model. 
The innovations of AI Flow provide enhanced intelligence, timely responsiveness, and ubiquitous accessibility to AI services, paving the way for the tighter fusion of AI techniques and communication systems.
These advancements are crucial to numerous application scenarios, including but not limited to embodied AI, wearable devices, and smart cities.



\end{abstract}




\section*{Keywords}
AI Flow, distributed inference, edge artificial intelligence, large language model, collective intelligence



%
\IEEEpeerreviewmaketitle

\section{Introduction}
%
%
%
%
The foundational breakthroughs of Alan Turing in artificial intelligence (AI), notably the conceptualization of machine intelligence and the Turing Test \cite{turing2009computing}, alongside the birth of information theory by Claude Shannon \cite{shannon1948mathematical}, stand as defining milestones laying the foundation of information technology (IT) and communication technology (CT), respectively. 
The development of IT and CT techniques follows a dual trajectory: enhancing the capabilities of individual machines and creating networks for better interconnection among multiple machines.
\textbf{Recently, the rapid advancements in AI have further propelled the fusion of IT and CT, establishing a symbiotic relationship between them: the IT-CT integration drives the transformative AI breakthroughs by providing exponentially growing data streams and computing infrastructure, while AI emerges as an active force reshaping their evolutions~\cite{ict_survey}. }
This dynamic interplay reached a watershed moment in 2022 when ChatGPT, a large language model (LLM) developed by OpenAI, captured global attention and sparked a wave of innovation across various sectors, ushering in a new era of digital transformation.

Today, the emergence of large models ushers in a new era of AI. Defined by their capabilities to generate high-quality multimodal content and comprehend open-ended questions, large models have demonstrated remarkable performance in natural language processing, computer vision, and other domains~\cite{tom2020_llm}. These unprecedented capabilities have fueled aspirations to bring the power of advanced AI technologies to the fingertips, which coincidentally aligns with the vision of ubiquitous intelligence in future communication networks~\cite{6gvisions}. 
However, the rapid expansion in the size of AI models has triggered a surge in the demand for hardware capacities such as computation, storage, and power. This trend creates substantial barriers to deploying large-scale AI models on heterogeneous devices, especially resource-constrained devices such as internet-of-thing (IoT) sensors and mobile phones. Although prior studies have proposed to exploit the idle resources across distributed platforms, e.g., high-performance servers at the cloud and edge servers at the base stations, via communication networks to alleviate hardware limitations, the transmission of high-dimensional data exacerbates the communication overhead, posing a critical bottleneck for modern communication infrastructure.
\textbf{Thus, achieving the vision of ubiquitous intelligence urges multidisciplinary breakthroughs at the intersection of AI and communication technologies. 
This requires a broad consensus on standardized frameworks and protocols.}

\begin{figure}[t]
    \centering
    \includegraphics[width=\linewidth]{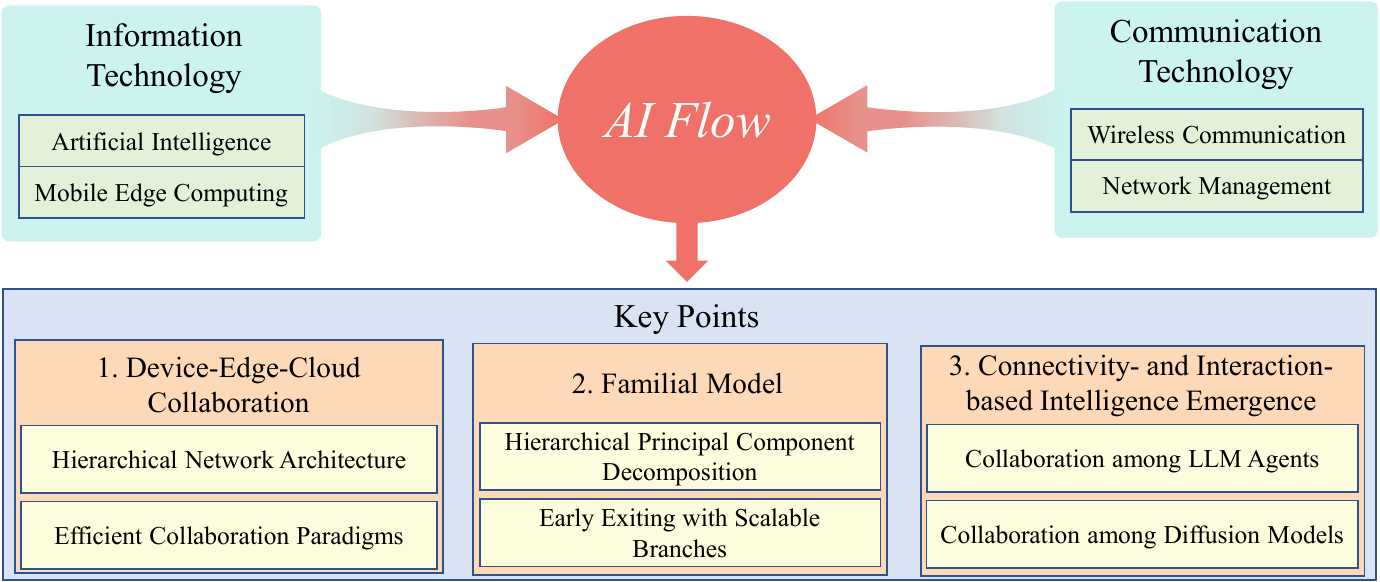}
    \caption{An overview of the AI Flow framework.}
    \label{fig_aiflow}
\end{figure}

In this paper, we introduce the novel concept of AI Flow and its underpinning technologies, which represent a transformative paradigm at the intersection of AI and communication networks~\cite{jiawei2025_aiflow,mao2017survey,shao2021communication,letaief2019roadmap,letaief2021edge}. As illustrated in Fig.~\ref{fig_aiflow}, AI Flow is a multidisciplinary framework designed to enable seamless transmission and emergence of intelligence across hierarchical network architectures by leveraging inter-agent connections and human-agent interactions. At its core, AI Flow emphasizes three key points: 
\begin{itemize}
    \item \textit{Device-Edge-Cloud Collaboration:} 
    \textbf{AI Flow leverages a unified device-edge-cloud architecture, integrating end devices, edge servers, and cloud clusters, to dynamically optimize scalability and enable low-latency inference of AI models. }
    By developing efficient collaboration paradigms tailored for the hierarchical network architecture, the system minimizes communication bottlenecks and streamlines inference execution.
    \item \textit{Familial Models}: Familial models refer to a set of multi-scale architectures designed to address diverse tasks and resource constraints within the AI Flow framework. 
    \textbf{These models facilitate seamless knowledge transfer and collaborative intelligence across the system through their interconnected capabilities. Notably, the familial models are feature-aligned, which allows efficient information sharing without the need for additional middleware. }
    Furthermore, through well-structured collaborative design, deploying familial models over the hierarchical network can achieve enhanced inference efficiency under constrained communication bandwidth and computational resources.
    \item \textit{Connectivity- and Interaction-based Intelligence Emergence}: AI Flow introduces a paradigm shift to facilitate collaborations among advanced AI models, e.g., LLMs, vision-language models (VLMs), and diffusion models, thereby stimulating emergent intelligence surpassing the capability of any single model. 
    \textbf{In this framework, the synergistic integration of efficient collaboration and dynamic interaction among models becomes a key boost to the capabilities of AI models.}
\end{itemize}

\subsection{The Rise of AI} 
The concept of AI, first founded as an academic discipline in 1956, has undergone transformative growth through iterative advancements. Early progress in AI centered on foundational machine learning (ML) theories, including regression analysis, support vector machines (SVMs), and artificial neural networks. Driven by the rapid development in computational power and data availability, breakthroughs in deep learning later revolutionized the field \cite{li1, li2, li3, li4, li5}. Pioneering models like AlexNet and ResNet attained unprecedented accuracies in image classification, object detection, and speech recognition, catalyzing their widespread adoption across industries such as healthcare, robotics, and autonomous systems. Reinforcement learning (RL) demonstrated its superhuman capabilities in domains requiring long-term planning as well. 

Driven by collaborative advancements in academic research and industrial innovation, the emergence of large generative models has revolutionized the AI domain over the past decade. 
Unlike traditional AI models that are confined to highly specialized tasks, generative AI marks a paradigm shift by enabling models to synthesize novel content, including text, images, and audio. Rather than focusing on task-specific optimizations, this paradigm shift leverages open-ended learning through massive-scale pretraining on diverse data, allowing it to transcend the rigid boundary of traditional AI.
The Transformer architecture~\cite{ashish2023_attention} stands out as one of the most pivotal breakthroughs in the evolution of LLMs and serves as the backbone for many state-of-the-art AI models. 
By employing the self-attention mechanism, a dynamic approach to evaluate the relationship between different elements of the input data, the Transformer effectively mitigates the limitation on modeling long-range dependency that hinders conventional architectures like recurrent neural networks (RNNs) and convolutional neural networks (CNNs). This architectural innovation has catalyzed the development of LLMs such as GPT-4~\cite{openai2024_gpt4}, Deepseek-V3~\cite{deepseekai2025_deepseekv3}, and Qwen-2.5~\cite{qwen2025_qwen2.5}, showcasing their versatility and task generalization abilities. Therefore, LLMs have attracted immense popularity and have risen to prominence as indispensable tools across manifold domains (e.g., healthcare, education, and finance), shining as a key milestone of AI.

To equip the text-based LLMs with multimodal capabilities, large VLMs integrate visual and linguistic reasoning by harnessing the prior knowledge obtained from the large-scale pretrained vision encoders~\cite{haotian2023_llava}. Typically, VLMs comprise two primary components: a vision encoder and an LLM.
The vision encoder is responsible for extracting hierarchical visual features from raw pixel data. These features are then aligned with the embedding space of the LLM via a lightweight adapter. 
As such, the visual information captured by the vision encoder can be effectively transferred to the LLM, bridging the gap between visual and textual modalities. 
Despite the success in content generation and visual understanding, their adoption in complicated real-world applications such as logical reasoning, collective decision-making, and multi-agent collaboration remains limited.

To address the aforementioned limitation, a series of innovative techniques have been proposed to enhance the capabilities of AI models~\cite{li2022positive}. 
One of the most notable techniques is the incorporation of human-like reasoning mechanisms. 
The chain-of-thought (CoT) framework~\cite{jason2023_cot}, which guides an AI model to produce intermediate reasoning steps before arriving at the final answer, has established itself as a leading solution for intricate problems.
Besides, RL has also proven instrumental in aligning model outputs with human preferences. The synergy of CoT and RL has unlocked a new pathway for advancing the capabilities of LLMs.
Moreover, collective intelligence is a widespread phenomenon where individuals exhibit simple behaviors that lead to complex group dynamics. 
The emergent intelligence from these interactions has inspired the development of LLM-based multi-agent systems (MASs)~\cite{mas_2024,jiawei_wirelessllm}, which are designed to resolve complex real-world problems based on multi-stage reasoning among agents. Specifically, heterogeneous agents with distinct capabilities aim at achieving a common goal through adaptive collaboration via a distributed network, thereby attaining intellectual capacity that outperforms that of the individual agents. 
These remarkable advancements in model capability have directly enabled a surge of emerging applications in numerous fields. 
A key trend is to provide ubiquitous intelligence services at the network edge, thereby bringing intelligence closer to users.


\subsection{Towards Ubiquitous AI Applications}
\textbf{The vision of ubiquitous intelligence targets enabling AI services with unparalleled capabilities across various sectors by pushing AI to the network edge~\cite{shao2023task,mao2017survey,letaief2019roadmap}}, where AI is seamlessly integrated into everyday environments, thereby empowering both end devices and edge servers to deliver high-quality services.
This shift unlocks unprecedented capabilities of AI across diverse sectors, including smart cities, industrial automation, and autonomous systems, by ensuring AI services are both responsive and adaptable to dynamic environments.
For instance, AI-powered glasses augment human perception by highlighting critical visual features, providing navigational guidance, or overlaying relevant contextual information about the environment, thereby significantly enhancing spatial awareness. 
Furthermore, devices integrated with LLMs have the potential to enable seamless cross-linguistic communication through real-time spoken language translation. However, the demanding requirements of sensing and the processing of high-volume, multimodal data streams present significant challenges for resource-constrained platforms like headsets. Addressing these limitations necessitates the development of more efficient on-device processing techniques and optimized data transmission strategies at the network edge.

\textbf{Collaborative MAS at the network edge represents a groundbreaking paradigm where interconnected intelligent agents across the network dynamically collaborate to accomplish sophisticated tasks requiring real-time coordination and adaptability in real-world scenarios.}
A prominent direction in this domain is embodied AI, where autonomous agents such as drones and robots physically interact with dynamic environments to perform tasks like disaster response. To this end, efficient collaboration between heterogeneous agents with complementary capabilities is favorable, leading to the necessity of seamless information sharing and decentralized control. Specifically, these agents must jointly negotiate operational roles, fuse multimodal sensory data, and adapt to evolving environmental conditions.
However, achieving such collaborative intelligence remains challenging due to the dynamic network conditions and limited on-board computational capacities of agents, which collectively impede latency-sensitive decision-making processes. Despite the potential of collaborative MASs to bring about revolutionary capabilities, realizing the vision of ubiquitous AI applications hinges on overcoming substantial obstacles inherent to resource-constrained environments.


\subsection{Challenges}
Although the future of ubiquitous AI applications is promising, a critical question comes: \textit{how far away exactly are we from realizing this vision?} To answer this question, we present a comprehensive analysis of the current landscape. 
The deployment of emerging AI applications confronts significant challenges stemming from hardware resource limitations and communication network constraints, which collectively hinder the scalability, responsiveness, and sustainability of real-world systems.

\subsubsection{Hardware Resource Limitations}
The evolution of AI has been marked by the exponential growth in model scale and computational complexity, directly escalating hardware requirements for AI deployment. For instance, early AI models like ResNet~\cite{resnet2016} ($11.69 \sim 60.19$ million parameters) proposed in 2016, could be efficiently executed on consumer-grade hardware, whereas modern LLMs are operating at scales spanning billions to trillions of parameters, e.g., Llama-4 ($0.1 \sim 2$ trillion parameters) released in 2025. This over a hundredfold growth in model scale in less than a decade hinders the deployment of LLMs on common devices. Memory bottleneck emerges as a critical challenge, particularly in mobile systems with limited storage capacity. For instance, inference-phase memory footprints can surge to tens to hundreds of gigabytes (GBs) with advanced large models, far exceeding the capacity of typical end devices ($\sim $4-32 GBs). Concurrently, the computational complexity of an AI model escalates with its scale, undermining latency-sensitive applications like autonomous driving on low-end devices. Moreover, the incorporation of human-like reasoning mechanisms in LLMs further exacerbates computational burdens, primarily accounting for the generation of intermediate reasoning tokens. 
Since the computational complexity of LLMs scales quadratically with sequence length on account of the self-attention mechanism, the inclusion of these reasoning tokens elevates the overall computational cost during inference.

Extensive research has focused on developing efficient LLMs to accommodate consumer-grade hardware. However, they generally yield efficiency gains by sacrificing model capabilities, thereby compromising the reliability of the deployed system. 
Techniques such as quantization, which reduces model precision to decrease memory footprint, and model pruning, which removes redundant parameters, are prominent examples. These limitations highlight the urgent need for novel mechanisms that reconcile scalability with hardware-aware design principles, ensuring the practical deployment of LLMs across heterogeneous platforms without sacrificing functional integrity.

\subsubsection{Communication Network Constraints}
The realization of ubiquitous intelligence also faces significant challenges posed by the inherent limitations of the communication network. The inherent vulnerability of typical wireless communication networks poses a substantial challenge in ensuring robust connectivity and efficient data transmission across heterogeneous environments, particularly in scenarios characterized by high mobility (e.g., vehicular networks) or ultra-dense user connections (e.g., smart cities). This vulnerability critically undercuts the reliability and security that are essential for ubiquitous intelligence systems. 
Moreover, the instability of core network infrastructure also introduces critical reliability challenges. Specifically, network congestion, often exacerbated by excessive traffic demand that overwhelms the infrastructure capacity, can lead to packet loss and queuing delays, impairing the reliability and efficiency of cloud resource access. Additionally, network jitter, arising from the dynamic network conditions and routing changes, can introduce unpredictable latency, which adversely affects the performance of real-time applications.

The communication overhead escalates significantly in distributed and collaborative inference scenarios due to the frequent exchange of high-dimensional data, thereby straining the network bandwidth resources and introducing latency bottlenecks. Specifically, in split-inference frameworks, end devices must transmit activation features to the edge servers for subsequent processing, often generating tens to hundreds of megabytes of data per inference step. The cumulative data transfer can overwhelm the network infrastructure, leading to increased latency and reduced throughput. Similarly, the iterative synchronization of reasoning traces among agents in LLM-based MASs further amplifies the communication overhead and hinders the scalability of the systems. 

\textbf{In a nutshell, ubiquitous AI confronts a dual-bottleneck challenge spanning hardware resources and communication networks, thus demanding multidisciplinary solutions to harmonize the deployment of large models across heterogeneous hardware platforms.} To this end, \textit{AI Flow} emerges as a promising framework, which streamlines the inference processes of models by leveraging hierarchical network architectures and unifying techniques from both AI and communications domains. With holistic optimizations of heterogeneous network resources and strategic co-designs of the interaction protocols for familial models, AI Flow facilitates scalable and efficient deployment of the models, thus attaining real-time response and cross-platform operations essential for AI applications. In the next subsection, we will introduce the potential solutions for bridging the aforementioned challenges under the AI Flow paradigm.

\begin{figure}[ht]
    \centering
    \includegraphics[width=\linewidth]{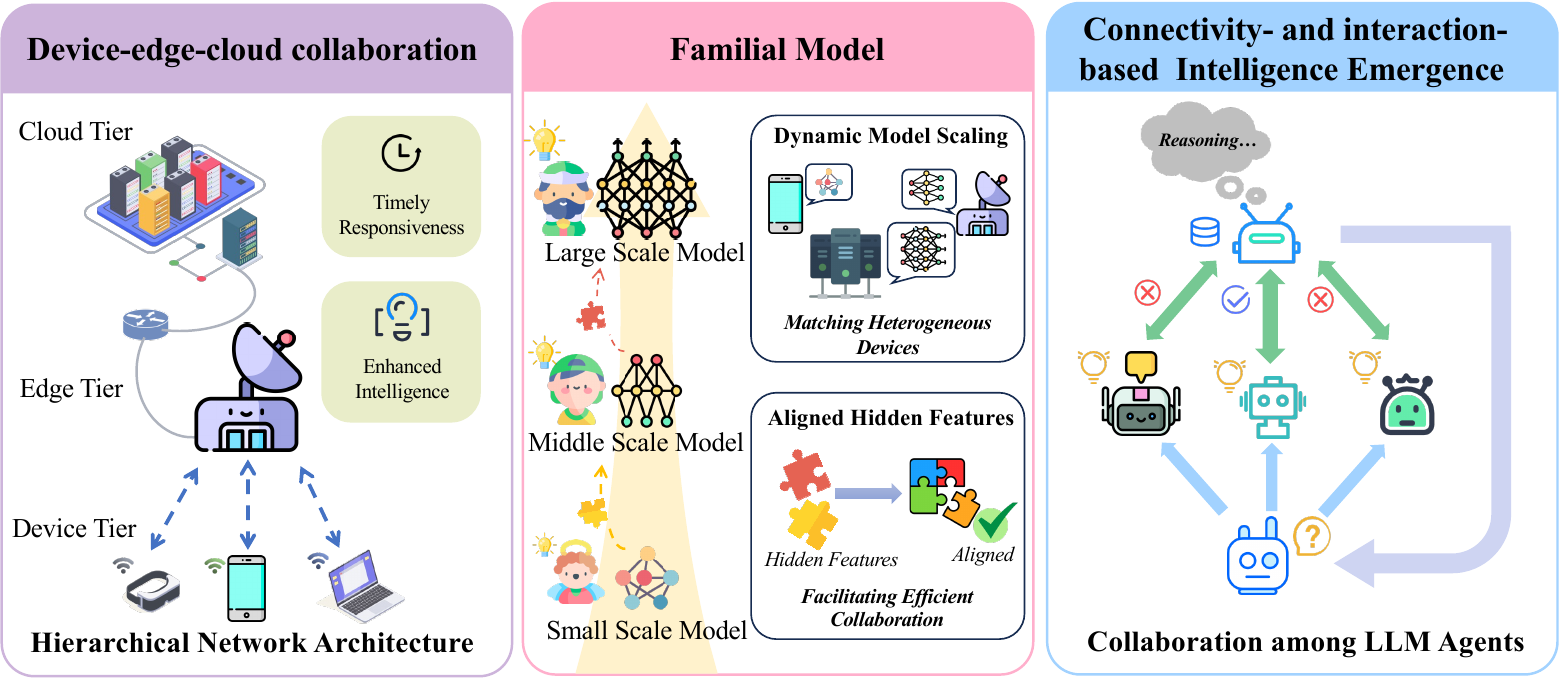}
    \caption{A comprehensive framework of AI Flow highlighting its three key points.}
    \label{fig2}
\end{figure}

\subsection{Addressing Challenges with AI Flow}
The convergence of the aforementioned challenges in modern AI systems necessitates multidisciplinary solutions to harmonize the deployment of large models across heterogeneous hardware platforms. 
To address these challenges, we develop the concept of AI Flow, which streamlines the inference processes of AI models, as illustrated in Fig.~\ref{fig2}. In particular, by leveraging the device-edge-cloud architecture, AI Flow introduces a suite of techniques targeted at enabling efficient collaboration across distributed resources, which effectively mitigate hardware constraints and reduce high inference latency.
To fully exploit the potential of the hierarchical network, there is a critical need to develop models that are specifically tailored to heterogeneous resources and diverse downstream tasks. Toward this end, we propose the development of familial models that can achieve optimized inference efficiency under resource-constrained conditions. Furthermore, AI Flow facilitates interaction and interconnection among agents, fostering intelligence emergence and achieving superior performance compared to single-model systems.


\subsubsection{Device-Edge-Cloud Collaboration}
Serving as the underpinning foundation of AI Flow, \textbf{device-edge-cloud collaboration optimizes resource utilization and latency-accuracy trade-offs by orchestrating AI workflows across heterogeneous hardware platforms, arising as a favorable solution for handling the stringent hardware constraints~\cite{yifei_edgeai}. }

Modern communication networks can be modeled as a three-tiered architecture to efficiently distribute computational tasks based on hardware capabilities and latency requirements. At the lowest tier, end devices, typically characterized by limited processing and storage resources, are best suited for executing lightweight tasks and real-time interactions. The intermediate tier is comprised of edge servers that offer modest computational capabilities but are located in close proximity to the user. This proximity enables the edge servers to minimize data transmission delay, allowing them to provide complementary computing resources to end devices and prioritize expedited processing for latency-sensitive applications. The top tier, represented by cloud clusters, delivers high-capacity, scalable computing resources designed to handle computation-intensive workloads such as large-scale model training. 

The integration of remote cloud clusters, distributed edge servers, and resource-constrained end devices has emerged as a promising solution for addressing hardware resource limitations. This hierarchical architecture enables AI Flow to dynamically adapt to heterogeneous hardware capabilities and application requirements, meeting the multifaceted requirements of AI applications. Specifically, the three-tier network infrastructure is designed to balance resource scalability and low-latency inference for supporting various models, thereby alleviating the hardware resource bottlenecks due to the dramatic growth in model scales. Through hierarchical model deployment and task offloading, latency-critical inference is prioritized at edge tiers and device tiers while cloud resources are leveraged for resource-intensive operations. 
Despite the potential of the device-edge-cloud paradigm to address hardware limitations in realizing ubiquitous intelligence, persistent challenges in collaboration efficiency demand considerable improvement. Hence, several pivotal techniques are developed for the hierarchical network to achieve optimal system performance.

A task-oriented feature compression mechanism tailored for device-edge cooperative VLM inference is proposed under the framework of AI Flow. In device-edge co-inference scenarios, the split-inference framework typically suffers from prohibitive communication overhead due to the transmission of high-dimensional intermediate data, limiting their scalability and deployment in bandwidth-constrained environments. This necessitates communication-efficient designs to mitigate bandwidth consumption while maintaining inference accuracy. To this end, the proposed mechanism effectively alleviates the communication network constraints by integrating bandwidth-aware coding techniques and adaptive token pruning approaches to their design. This integration enables dynamic optimization of task-relevant data payloads in response to real-time channel conditions.

To further enhance the collaboration efficiency of the device-edge-cloud architecture, we propose a speculative decoding mechanism seamlessly embedded within the hierarchical network architecture. Specifically, models of varying scales are strategically deployed across network tiers in accordance with resource availability. This mechanism utilizes lightweight device models to predict the candidate tokens of LLMs, while larger edge and cloud models validate and refine these predictions. Through this paradigm, the inference process can be accelerated significantly by shifting the role of larger models from full sequence generation to error correction. Moreover, it is capable of attaining an adaptive quality-efficiency tradeoff by dynamically selecting optimal-capability models for each verification stage based on real-time resource constraints.

\subsubsection{Familial Models}
\textbf{In order to further exploit the heterogeneous resources across the hierarchical network architecture, AI Flow devises a novel paradigm for obtaining familial models that scale flexibly to arbitrary model sizes.} Within this paradigm, two key techniques are introduced: early exit and weight decomposition. The inherent architectural scalability of familial models allows them to fully utilize the heterogeneous hardware and meet diverse task demands, thereby enabling efficient collaboration among familial models, even under stringent resource constraints. 

Early exit leverages the pretrained intermediate representations, enabling models to terminate at earlier layers while maintaining acceptable performance. Through coarse-grained splitting of model execution, smaller variants achieve faster inference with minimal accuracy loss. A salient benefit of this approach is that it ensures the intermediate features are aligned across the familial models, facilitating the direct reuse of intermediate results from smaller variants without any middleware. The elimination of computational redundancy facilitates efficient deployment in heterogeneous networks.

To enable finer-grained scalability, weight decomposition is proposed as the complementary solution to early exit. In particular, weight decomposition enables decomposing model weights into low-rank matrices, yielding models with arbitrary numbers of parameters while minimizing redundancy. Concurrently, tailored weight initialization schemes are introduced to reduce the distortion in intermediate feature representations during the decomposition process. By decoupling model capability from given structural priors, the framework guarantees seamless interoperability across heterogeneous scaling configurations without incurring severe performance degradation or computational overhead.

\subsubsection{Connectivity- and Interaction-based Intelligence Emergence}
\textbf{Information capacity represents the ratio of information volume to data volume, representing the information-carrying capacity per unit data volume. 
To enhance the information utilization efficiency of agents, AI Flow reinforces connectivity and interaction among LLMs, VLMs, and diffusion models through the hierarchical network architecture.}
Specifically, a device-server collaboration scheme is proposed to improve the coherence and accuracy of the final responses generated by LLMs and VLMs, which follows a four-step pipeline.
First, the powerful server model identifies the most suitable devices according to the user query and descriptions of available device models and sends inference requests to the selected devices.
Second, devices generate responses using their own models optimized for specific domains or user preferences and forward them to the server, providing diverse insights from their perspectives.
Third, the central server model aggregates these intermediate results to synthesize a unified answer to the user query.
Fourth, the aggregated answer is returned to devices for possible revision, incorporating its specialized opinion with the global context.

In addition, Al Flow provides paradigms for collaboration among diffusion models, including serial, parallel, and networked collaboration schemes, tailored for tasks such as motion generation, depth estimation, and multimodal video generation.
To generate complex and coordinated multi-person motion sequences, we introduce a serial collaboration paradigm.
This method first composes temporally controlled motion segments with one model, in which the interaction between people is then refined by another model to improve coherence.
For the challenging monocular depth estimation task, we employ a parallel collaboration paradigm, beginning with extracting intermediate features with a unified image encoder.
Subsequently, two separate decoders are employed in parallel to estimate near and far regions, respectively.
Results from two decoders are fused based on a predicted anchor depth to obtain a complete and accurate depth map.
Another formidable problem for diffusion models lies in maintaining consistency and coherence in both spatial and temporal dimensions during video generation.
To address this limitation, we formulate a hierarchical networked collaboration paradigm to aggregate information across diverse modalities, including RGB, segmentation, depth, and audio, using separate encoders possibly deployed on distributed hardware.
Afterward, a unified diffusion transformer collectively processes these features, capturing correlations among different modalities to enforce spatial and temporal alignment.
Finally, the output in each modality is produced by a specialized decoding head according to the requirements of end users.

\subsection{Paper Organization}
The rest of the paper is organized as follows: Section~\ref{sec:key1} introduces the underlying device-edge-cloud architecture and outlines two collaborative inference paradigms within the framework of AI Flow. Section~\ref{sec:key2} details the familial model framework and its implementation strategies. Section~\ref{sec:key3} explores the synergistic capabilities of model collaboration, revealing the intrinsic pathways through which collective intelligence emerges. Section~\ref{sec:app} highlights the emerging applications and demonstrates the advantages of AI Flow in tackling the fundamental challenges within these domains. Finally, Section~\ref{sec:future} discusses future directions for expanding its scope, and Section~\ref{sec:conclusion} concludes the paper.

\section{Device-Edge-Cloud Collaboration}
\label{sec:key1}
\textbf{AI Flow leverages a hierarchical device-edge-cloud architecture to achieve both enhanced intelligence and timely responsiveness for AI services.
This three-tiered network architecture enables a flexible distributed inference pipeline for diverse downstream tasks, serving as the foundation of model collaboration, a cornerstone of AI Flow.}
In the following subsections, we first elaborate on the three network tiers, constituted by end devices, edge servers, and cloud clusters.
Subsequently, we formulate two collaboration paradigms across these tiers that highlight the inference efficiency of AI Flow.
First, we propose a task-oriented feature compression mechanism for device-edge cooperative VLM inference to minimize data transmission overhead and end-to-end system latency.
Second, we introduce a hierarchical collaborative decoding technique across multiple network layers to accelerate LLM inference while maintaining model intelligence.

\subsection{Hierarchical Network Architecture}
Typically, existing communication networks can be represented by a three-tier architecture according to the characteristics of the hardware platforms, which is detailed as follows:
\begin{itemize}
    \item \textit{Device Tier:} The device tier encompasses resource-constrained devices (e.g., smartphones, IoT devices, and laptops) that interface directly with users or environments. 
    Serving as the primary interface for data acquisition and initial processing, these devices critically influence the responsiveness and efficacy of the AI Flow framework. However, their operational scope is constrained by limited computational capacities, restricting them to deploying lightweight models and executing basic tasks such as local data aggregation, data preprocessing, and simple inference. Hence, the devices at this tier necessitate efficient collaboration with other tiers to augment their intellectual capacities.
    \item \textit{Edge Tier:} In this tier, edge servers are usually deployed at the network edge, e.g., base stations (BSs) and roadside units (RSUs), providing moderate computational capacities with relatively low transmission latency due to their proximity to end devices. The edge servers act as intermediaries between the cloud and device tiers, enabling localized processing and dynamic task orchestration. By taking over latency-sensitive workloads from resource-constrained end devices, the edge tier enhances responsiveness while reducing reliance on distant cloud infrastructure. However, its hardware resources remain limited compared to cloud clusters. Therefore, edge servers are imperative for the dynamic orchestration of workload by offloading exceptionally high compute-intensive operations to the cloud tier while directly supporting end-tier devices, ensuring the efficient utilization of hierarchical resources.
    \item \textit{Cloud Tier:} Cloud clusters of this tier are equipped with high-performance hardware resources (e.g., large storage capacities, high-grade CPUs/GPUs) and are interconnected via high-bandwidth links, thus capable of handling extremely compute-intensive tasks like large-scale model training. As the computational powerhouse of the hierarchical network architecture, this tier provides scalable resources to address the most demanding workloads. However, its physical distance from end devices introduces the highest transmission latency among the three tiers, rendering it unsuitable for latency-critical operations. To optimize hierarchical resource utilization, the cloud tier collaborates with lower tiers by offloading latency-insensitive tasks from device and edge tiers, while aggregating insights from distributed nodes to refine global models. This interplay ensures efficient allocation of workloads, leveraging the advantages of cloud tier in computational scale while mitigating its latency drawbacks through strategic task orchestration.
\end{itemize}

Driven by the availability of abundant computational resources in centralized data centers, cloud-centric systems have dominated the burgeoning development of AI over the past two decades. 
These systems excel in addressing resource-intensive challenges, especially in training models and large-scale model inference. 
However, such systems are often plagued by high latency and network volatility, hindering their deployment for latency-sensitive applications. 
These limitations have spurred the rise of edge AI systems, which exploit the distributed processing capabilities of endpoint devices and edge servers to facilitate low-latency execution. Nevertheless, this paradigm is not without its challenges, as it grapples with the constraints of limited computational capacities. 
Given the inherent resource demands of advanced AI models, edge AI systems may fail to meet the prospect of ubiquitous intelligence.

To bridge this gap, the synergistic integration over the hierarchical network presents a viable pathway forward. \textbf{By organizing computing and storage into a hierarchical framework, AI workloads can be dynamically orchestrated across heterogeneous hardware, i.e., end devices, edge servers, and cloud clusters. This adaptability ensures that applications meet strict latency, throughput, and scalability demands, unlocking the full potential of pervasive AI services.}

\subsection{Efficient Collaboration Paradigms}

\subsubsection{Task-Oriented Feature Compression}
\begin{figure}[t]
    \centering
    \includegraphics[width=.67\linewidth]{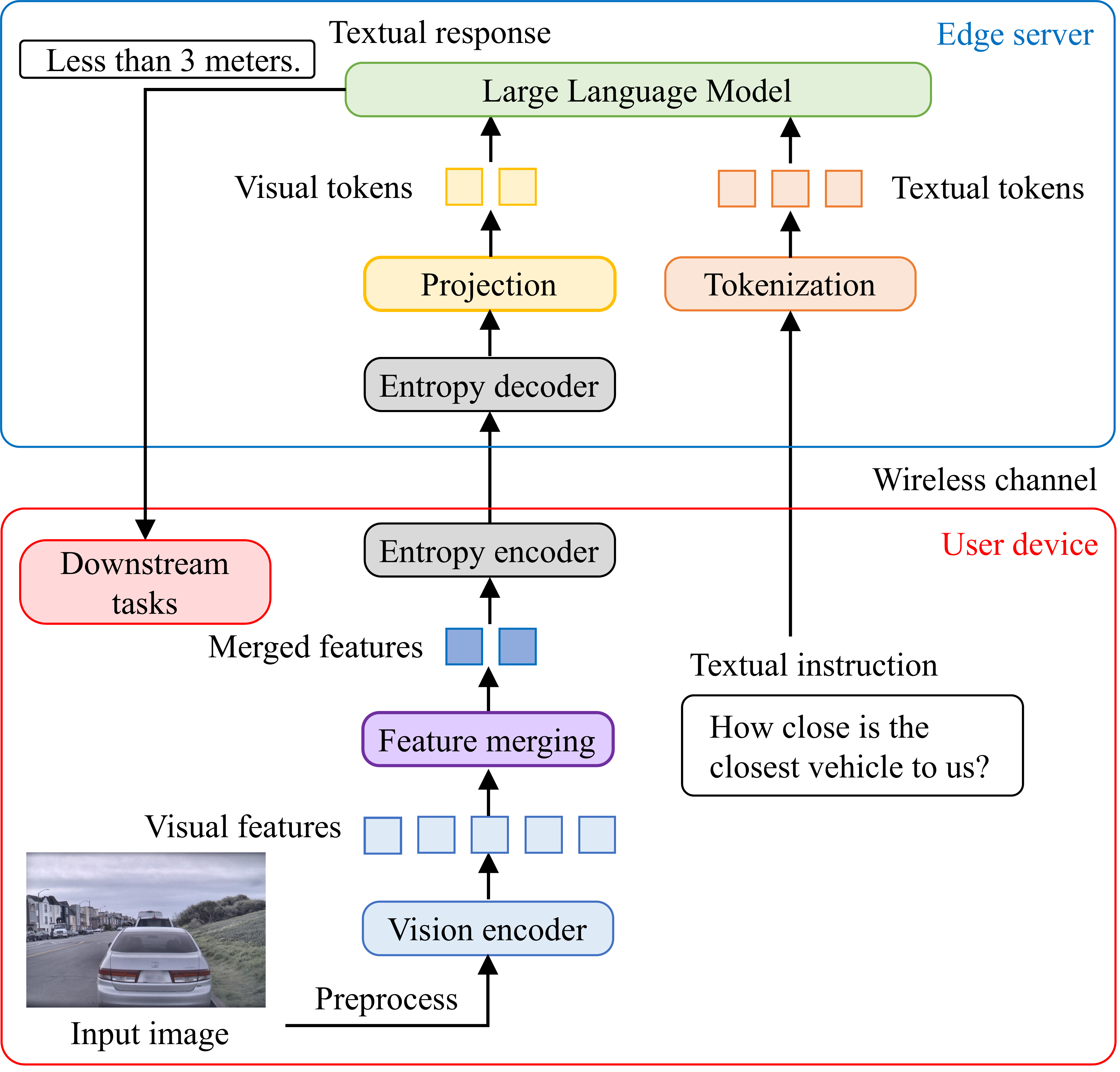}
    \caption{System diagram of the proposed TOFC method for device-edge co-inference.}
    \label{fig_TOFC_sys}
\end{figure}
With the rapid advancement of the multimodal understanding and reasoning capabilities of VLMs, VLM inference requests have soared at the network edge.
Due to the constraints on hardware resources of edge devices, the mainstream VLM inference pipeline involves transmitting the raw visual data to the cloud server, introducing prohibitive latency under limited uplink bandwidth.
Moreover, CLIP-based vision encoders \cite{clip_2021} generate a large number of visual tokens, which incurs excessive computational complexity for LLM inference and increases end-to-end system latency.
These two factors hinder the application of VLM in delay-sensitive tasks on edge devices, which highlights the necessity of reductions in both data transmission and inference time.
To address these limitations, \textbf{we propose a task-oriented feature compression (TOFC) method for device-edge co-inference \cite{TOFC,shao2022task}, which dynamically optimizes task-relevant visual feature transmission according to channel conditions via on-device merging and compression. }
Specifically, as illustrated in Fig. \ref{fig_TOFC_sys}, we first perform density peaks clustering based on $K$ nearest neighbors (DPC-KNN) \cite{DPC-KNN} on visual features generated by CLIP vision encoder, thus achieving substantial reductions in data size and computational load.
Subsequently, a hyperprior-based entropy model \cite{Balle2018} is employed to encode and decode merged features, thereby minimizing data transmission while maintaining performance on downstream tasks.
Finally, we train multiple entropy models specialized in encoding distinct features and adaptively select the optimal entropy model according to the characteristics of input features.

\textbf{Details.}
We choose cluster centers based on two criteria: the local density and the distance to other high-density features.
For each input feature, we first calculate the local density from the distances to its $K$ nearest neighbors.
Subsequently, the minimum distance to other higher-density features is computed, and the highest-density feature is assigned the maximum distance to allow it to be chosen as a cluster center.
We select features with the largest product of the local density and the distance to other high-density features as the cluster centers, while other features are grouped to the closest center.
Based on the clustering results, the visual features are merged through average pooling.

As entropy coding requires the probability distribution of the input variable, we choose the Laplacian distribution as the probability model of the merged features based on the measured probability density function (PDF).
The analysis network extracts the hyperprior from the merged features, and the synthesis network estimates the statistics of the merged features from the quantized hyperprior.
Considering the additive uniform noise introduced to the merged features during quantization, the conditional probability distribution of quantized features given the quantized hyperprior is modeled as a continuous relaxation of the Laplacian distribution, whose mean and scale are derived from the quantized hyperprior \cite{Balle2018}. 
We replace non-differentiable quantization with a straight-through estimator (STE) \cite{minnen2020channel} during training to allow back-propagation of gradients.

To improve the encoding efficiency, we employ a router network to rate all entropy models for each merged feature.
During inference, only the entropy model with the highest score is employed to encode and decode the merged feature.
During training, the output features are the weighted sum of the decoding results from all entropy models.
The weighting coefficient for each entropy model is calculated using the softmax function with temperature on the score vector.
The loss function is comprised of three components: the distortion loss, the rate loss, and the balance loss.
The distortion loss evaluates the performance of VLM inference, namely the cross-entropy loss calculated from the next token predicted by the LLM.
The rate loss is the combined entropy of quantized merged features and hyperprior, calculated from the estimated PDF.
The balance loss is introduced to encourage relatively balanced usage of all entropy models, formulated as the summed squared difference between the probability distribution of entropy model selection and random selection.

\begin{figure}[!t]
    \centering
    \subfloat[]{\includegraphics[width=.5\linewidth]{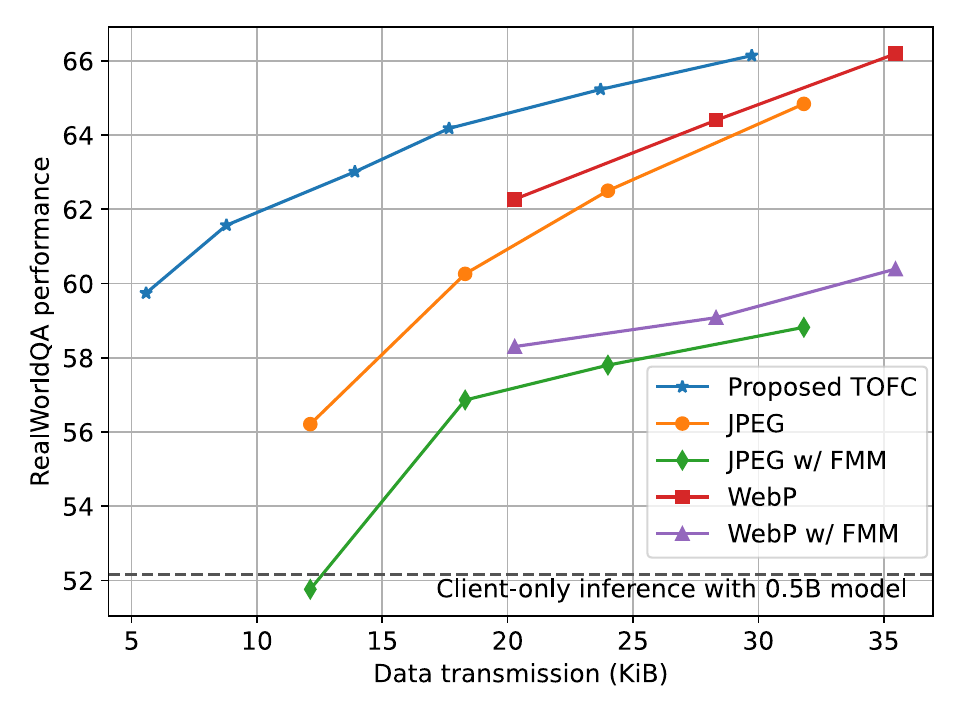}}
    \subfloat[]{\includegraphics[width=.5\linewidth]{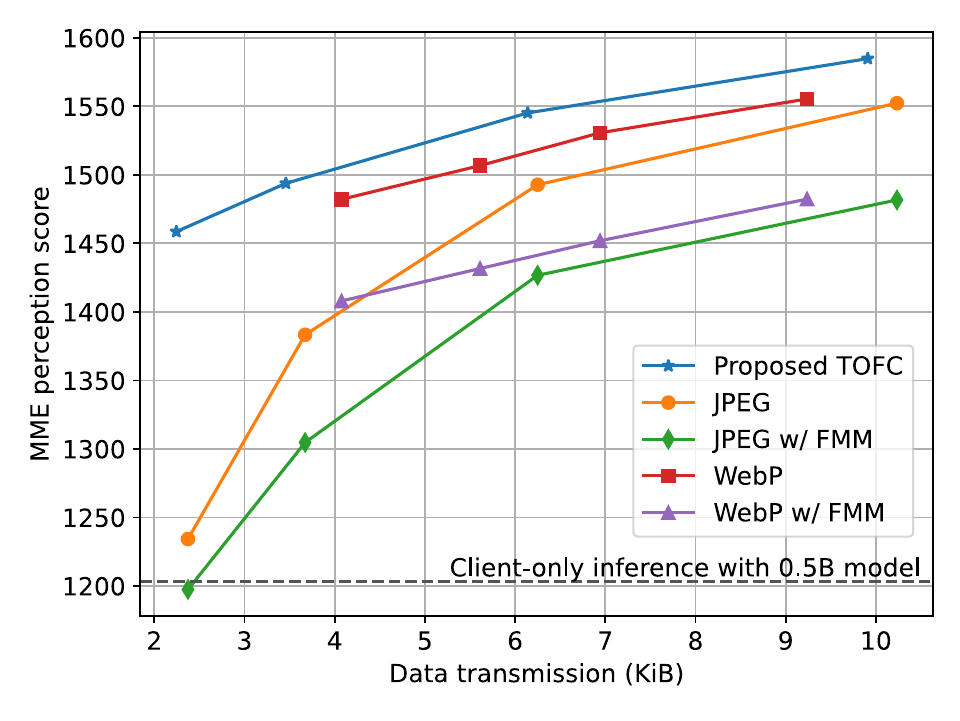}}
    \caption{The rate-performance curves of different inference methods in VQA benchmarks. Subfigures (a) and (b) show results on RealWorldQA and MME benchmarks, respectively. (w/ FMM: with the proposed feature merging module.)}
    \label{fig_TOFC_RD}
\end{figure}

\textbf{Evaluations.}
We conduct extensive experiments on the LLaVA-OneVision-7B model \cite{llava-ov}.
The ViT and LLM are fine-tuned using LoRA \cite{LoRA} with a rank of 64, and the multilayer perceptron (MLP) projector is full-parameter trained.
The model is trained for 1 epoch with the instruction tuning data used in the second training stage of LLaVA-1.5~\cite{llava1.5}.
The edge device is an NVIDIA Jetson AGX Orin, and the edge server is equipped with NVIDIA RTX 4090 GPUs.
We evaluate the multimodal understanding capability by RealWorldQA and the perception score of MME~\cite{MME}, in comparison to two data-oriented image compression methods: JPEG \cite{jpeg2000} and WebP~\cite{webp}.

Fig.~\ref{fig_TOFC_RD} demonstrates the rate-performance curves of the proposed TOFC and conventional image compression methods on RealWorldQA and MME benchmarks.
The reductions in data transmission overhead achieved by TOFC range from 25\% to 45\% and 35\% to 60\% for identical RealWorldQA performance in comparison to WebP and JPEG, respectively.
When the proposed feature merging is employed to ensure similar system latency to that of TOFC, traditional WebP and JPEG compressions require 5.9$\times$ and 6.7$\times$ data size to achieve task performance comparable to that of TOFC, respectively.
The data size savings provided by TOFC are approximately 25\% and 42\% for identical MME perception score, compared to WebP and JPEG, respectively.
With the proposed feature merging for inference acceleration, the reductions are even more prominent, equivalent to 69\% and 72\% of the data size required by WebP and JPEG, respectively.
These results highlight the efficiency of the proposed TOFC method achieved by discarding redundant visual features while maintaining high task performance.

\begin{figure}[!t]
    \centering
    \subfloat[]{\includegraphics[width=.5\linewidth]{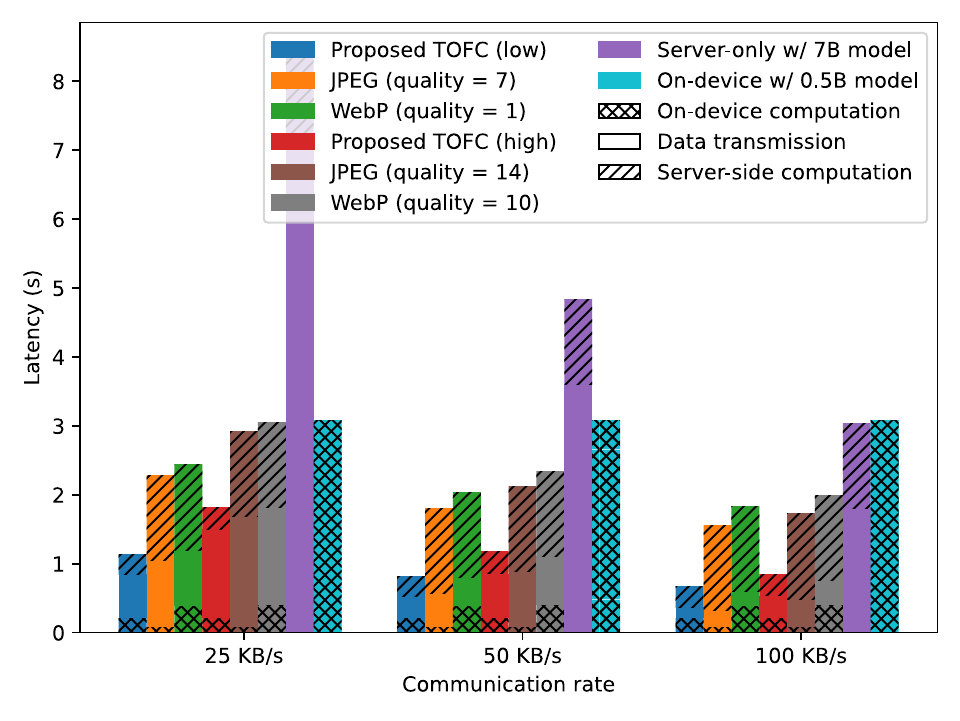}}
    \subfloat[]{\includegraphics[width=.5\linewidth]{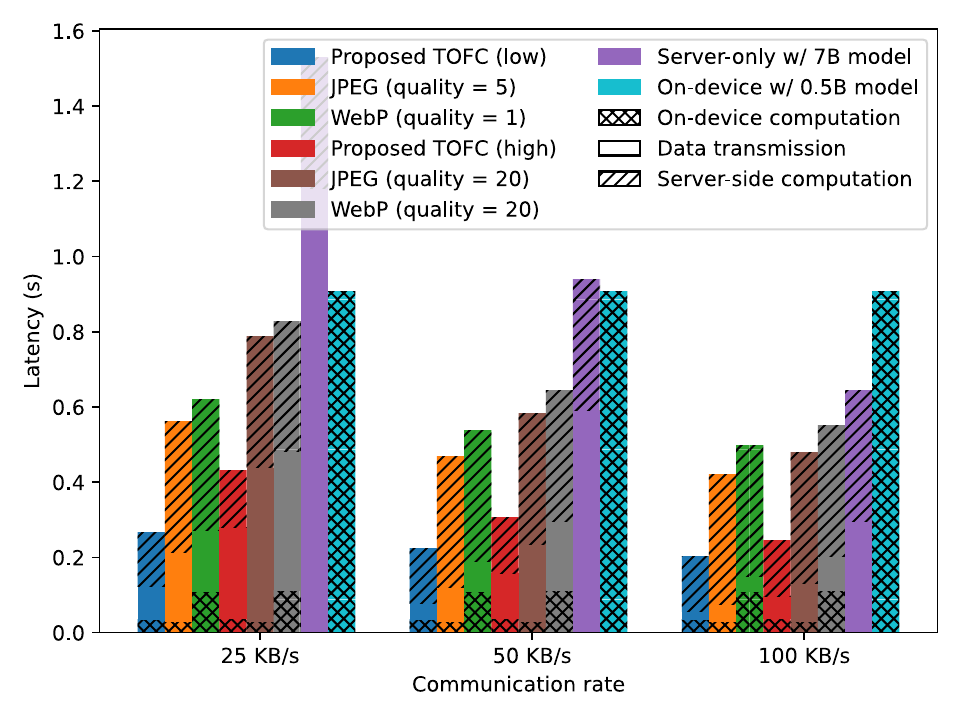}}
    \caption{Average inference latency per user request of different inference methods under different communication rates. The total latency consists of on-device computation time, data transmission latency, and server-side computation time. Subfigures (a) and (b) show results on RealWorldQA and MME benchmarks, respectively.}
    \label{fig_TOFC_latency}
\end{figure}
Fig. \ref{fig_TOFC_latency} illustrates the average inference latency per user request under different communication rates on the RealWorldQA and MME benchmarks.
The proposed TOFC method reduces system latency by half compared with WebP and JPEG for the RealWorldQA benchmark under low communication rates.
When the communication rates are higher, the latency reduction provided by TOFC is more substantial.
The inference latency of TOFC is only 32\% compared to that of server-only inference for the MME benchmark, with minimal performance degradation.
These results underscore the inference acceleration provided by the proposed TOFC method, thanks to significant reductions in the number of visual tokens.

\subsubsection{Hierarchical Collaborative Decoding}\label{sec: hierarchical collaboration decoding}
\begin{figure}[t]
	\centering
	\includegraphics[width=.98\linewidth]{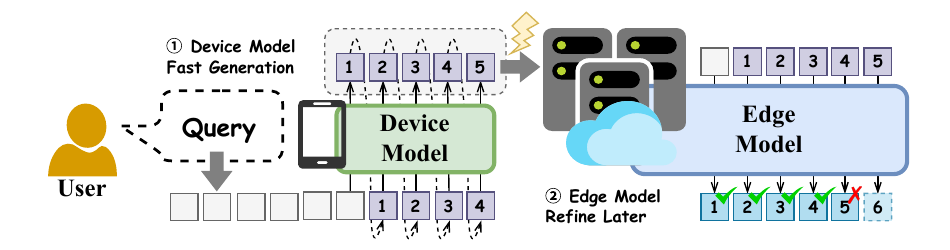}
	\caption{An overview of the hierarchical collaboration framework between devices and the edge server through speculative decoding.}
	\label{fig: speculative decoding framework}
\end{figure}

While edge devices can deliver efficient LLM services by transmitting compact input features through techniques like TOFC mentioned above, deploying fully-parameterized LLMs, which exhibit sufficient intelligence to address diverse downstream tasks, remains a formidable challenge in the resource-constrained environment.
Edge devices often face limitations in computational power, memory bandwidth, and energy efficiency, which hinder the execution of LLMs requiring billions of parameters to achieve state-of-the-art performance.
A promising solution to bridge this gap lies in speculative decoding, an efficient decoding paradigm that enables lightweight device models to emulate the capabilities of their server-based counterparts.
This approach operates by deploying a hierarchical collaboration framework.
A two-tiered case is shown in Fig. \ref{fig: speculative decoding framework}.
Specifically, a lightweight, device-optimized model generates draft responses or partial outputs locally, while a high-capacity edge model with more parameters acts as a verifier, refining the draft outputs through rapid validation and correction.
The interaction between the two models leverages the native intelligence on the device while preserving the high quality of responses generated by the powerful edge model.
In addition, speculative decoding minimizes the computational burden on edge devices by decoupling inference into a draft-then-verify workflow. 
The device model accelerates high-probability token predictions, while the edge model resolves ambiguities or complex reasoning tasks. 
This token-based synergy optimizes bandwidth usage, as only critical updates or corrections are transmitted between the user device and the edge server.
\textbf{In essence, such hierarchical device-edge collaboration redefines the boundaries of mainstream device-native AI capabilities, transforming resource-constrained devices into intelligent endpoints capable of delivering near-edge-side performance through strategic collaboration.}

\begin{figure}[t]
	\centering
	\includegraphics[width=0.98\linewidth]{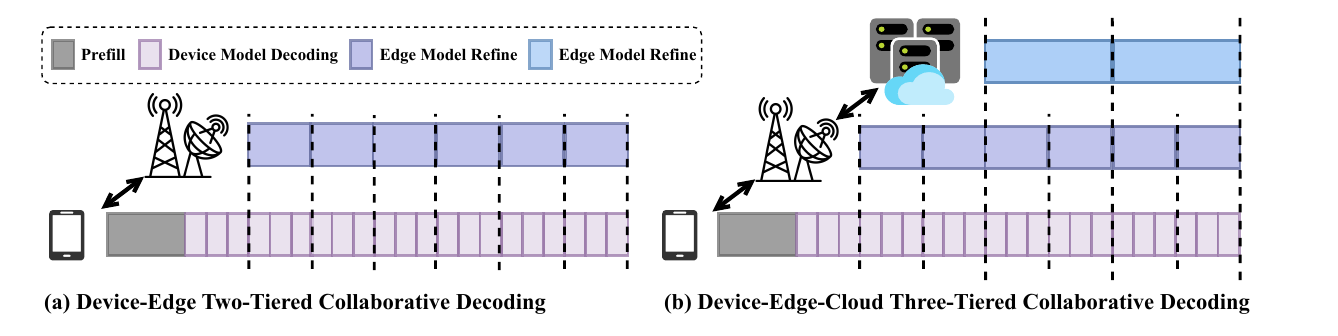}
	\caption{Proposed pipelines of hierarchical collaborative decoding.}
	\label{fig: speculative pipeline}
\end{figure}

\textbf{Details.}
We consider the device-edge scenario with two participants, where a lightweight model is deployed on the device, the source of user input, enabling fast and efficient on-device decoding.
Meanwhile, at the edge server, a more parameter-rich LLM participates in the decoding process, acting as a verifier that validates and refines the output generated by the device model.
Given the user instruction of length and a prefixed output sequence of length, the device model sequentially samples the next output token from the predicted probability distribution.
After generating several tokens locally, the entire predicted output sequence is transmitted to the edge server for validation.
At the edge, the model applies a soft acceptance strategy to evaluate each candidate token.
Specifically, a random seed is sampled from the uniform distribution.
For each candidate token, the edge model accepts it if the random seed is less than or equal to the ratio of the probability assigned by the edge model to that token, compared to the probability assigned by the device model.
This collaborative inference framework ensures that the lightweight device model can operate efficiently while still benefiting from the accuracy and reliability of the larger model deployed at the edge, achieving a balance between speed, resource usage, and output quality.

To overcome the inherent latency caused by the sequential draft-then-verify paradigm in traditional speculative decoding, we propose a parallel device-edge collaborative decoding framework.
The structure of this pipeline is illustrated in Fig. \ref{fig: speculative pipeline}, where a device model keeps generating tokens efficiently without blocking, while the edge server receives these tokens from the device at intervals, and refines them if necessary.
Here, in order to execute both device and edge computing with minimized blocking, we introduce a well-aligned inference pipeline, where the wall time for the device model to decode several tokens should be aligned with the time for the edge model to execute once forward pass on the received token sequence.
By combining the fast generation capability of the device model with the high intelligence of the edge model, our framework improves both the quality and generation efficiency of the final output. 
Moreover, the proposed collaborative architecture is highly flexible—it can be naturally extended to more complex setups such as a three-tiered device-edge-cloud architecture, allowing for even broader deployment scenarios and superior performance improvement.

\textbf{Evaluations.}
The accuracy and efficiency of the proposed device-edge-cloud collaborative decoding framework are validated through extensive experiments.
We conduct evaluations on Math-500 \cite{lightman2023lets} and AIME 2024 datasets for mathematical reasoning, as well as the LiveCodeBench \cite{jain2024livecodebench} dataset for code generation.
For all the benchmarks, we employ pass@1 as the performance metric.
For the two-tiered device-edge setup, we employ a two-model configuration: 1.5B (device) and 7B (edge).
In the three-tiered device-edge-cloud infrastructure, we extend this to a three-model configuration: 1.5B (device), 7B (edge), and 14B (cloud).
To establish performance baselines, we also report results from standalone decoding using each model individually.
All experiments were carried out on a system equipped with four NVIDIA H100-80GB GPUs.
Communication delay is considered negligible due to the small size of transmitted token indices, which results in minimal bandwidth consumption.

\begin{table}[t]
	\centering
	\caption{Comparison of different model configurations in diverse benchmarks.}
	\label{tab: device edge collaborative results}
	\begin{tabular}{lccccccc}
		\toprule
		\multirow{2.5}{*}{\textbf{Method}} & \multirow{2.5}{*}{\textbf{Model Size}} & \multicolumn{2}{c}{\textbf{MATH-500}} & \multicolumn{2}{c}{\textbf{LiveCodeBench}} & \multicolumn{2}{c}{\textbf{AIME 2024}}\\
		\cmidrule{3-4}\cmidrule{5-6}\cmidrule{7-8}
		& & \textbf{Acc} & \textbf{Speed} & \textbf{Acc} & \textbf{Speed} & \textbf{Acc} & \textbf{Speed}\\
		\midrule
		Device-Only & 1.5B & 83.9\% & 44.26 & 16.9\% & 43.51 & 28.9\% & 66.09\\
		Edge-Only & 7B & 92.8\% & 32.07 & 37.6\% & 35.42 & 55.5\% & 41.97\\
		Cloud-Only & 14B & 93.9\% & 27.86 & 53.1\% & 25.08 & 69.7\% & 36.31\\
		Device-Edge & (1.5B, 7B) & 92.8\% & 40.09 & 37.6\% & 39.98 & 55.5\% & 49.34\\
		Device-Edge-Cloud & (1.5B, 7B, 14B) & 93.9\% & 35.42 & 53.1\% & 33.98 & 69.7\% & 38.11\\
		\bottomrule
	\end{tabular}
\end{table}

Table \ref{tab: device edge collaborative results} presents a comparative analysis of both accuracy and efficiency between standalone decoding at each side and the proposed hierarchical collaborative decoding.
Both the two-tiered and three-tiered collaborative decoding frameworks exhibit notable improvements in response accuracy while preserving efficient decoding speeds that are faster than edge-only and cloud-only cases, respectively.
For instance, in the MATH-500 benchmark, device-edge collaborative decoding achieves around $40.09$ tokens per second generation, which is about $1.25\times$ speedup than edge model standalone decoding with the same accuracy.

\section{Familial Model}
\label{sec:key2}
\textbf{The familial model refers to a series of different-sized models with aligned hidden features, thereby enabling overhead-free information sharing and effective collaboration.
This architecture supports an almost arbitrary number of parameters, allowing it to fully exploit the computational capabilities of heterogeneous devices and satisfy the demands of various downstream tasks.}
Contrary to conventional models of varying sizes, familial models can directly reuse the intermediate results from smaller models without additional middleware.
Consequently, the cooperative inference of multiple familial models exhibits remarkable efficiency, particularly under stringent constraints on communication bandwidth and hardware resources.

In the following subsections, we first introduce related works and then elaborate on two enabling techniques of familial models: weight decomposition and early exit.
We propose two strategies to implement familial models for LLMs and VLMs based on the aforementioned enabling techniques.
Experimental results on familial models are presented to demonstrate their performance and adaptability.
Finally, discussions on the benefits and applications of familial models are given.

\subsection{Related Works}


\subsubsection{Weight Decomposition for Parameter Efficiency}
{
Low-rank decomposition is widely used to reduce model parameters while maintaining structural consistency.
Early work by Sainath et al. \cite{sainath2013lora} applied low-rank matrix factorization to speech models, while Yang et al. \cite{Yang_2020_CVPR_Workshops} explored training low-rank structures directly via SVD to preserve performance.
In this work, we develop upon these SVD-based approaches by introducing data whitening and truncating the components with the smallest singular values (Section \ref{layer_decomp}), enabling dynamic control of the decomposition rank per layer. This design ensures fine-grained parameter tuning while minimizing the distortion caused by approximating the original weight matrices.
Another line of approaches to enhance parameter efficiency by weight decomposition is LoRA, first proposed by Hu et al. \cite{LoRA}, which has become a widely adopted technique for efficiently fine-tuning large pre-trained models. It freezes the backbone parameters and inserts low-rank matrices for adaptation, achieving performance parity with significantly fewer trainable parameters.
Recent improvements, such as LoRA+ by Hayou et al. \cite{hayou2024loraefficientlowrank}, optimize learning rates and scaling for even larger models. In contrast to modern LoRA applications, our familial model architecture extends the concept to scalable branch networks, combining low-rank decomposed transformer blocks with shared LM heads to support efficient multi-branch inference and fine-grained tuning on the parameter count.
}

\subsubsection{Early-Exit Methods}
{
AI models typically make predictions by processing the input data through a series of structurally similar layers and exiting at the last layer.
To speed up inference and reduce computational costs, many studies \cite{rahmath2024early,chen2024eellm,kumar2025heliosadaptivemodelearlyexit} incorporate multiple exit points with side branches, allowing early exiting at an intermediate layer to reduce computational complexity.
Specifically, BranchyNet by Teerapittayanon et al. \cite{teerapittayanon2016branchynet} pioneered this idea by adding side branches to early layers. 
Branchy-GNN proposed by Shao et al. \cite{shao2021branchy} extended this approach to efficient point cloud processing based on graph neural networks.
Liu et al. \cite{liu2021anytime} proposed the first unified and end-to-end approach for anytime prediction, which attaches a cascade of exit points to make predictions from intermediate features.
More recent studies on employing early exit in transformer-based models include DeeBERT \cite{xin-etal-2020-deebert}, SkipBERT \cite{wang-etal-2022-skipbert}, and CALM \cite{NEURIPS2022_6fac9e31}.
Notably, EE-LLM \cite{chen2024eellm} implemented large-scale early-exit language models with 3D parallelism and KV cache compatibility. 
Pan et al. introduced a lightweight and economical solution to fine-tune early-exit LLMs~\cite{pan2024eetuningeconomicalscalablesolution}.
In this work, we propose a novel framework that introduces model decomposition to the branch networks after exit points. 
This design enables flexible parameter control while preserving alignment across branches, thereby supporting collaborative inference on heterogeneous devices. 
}



\subsection{Enabling Techniques}
\subsubsection{Weight Decomposition} \label{layer_decomp}
The first enabling technique of familial models is weight decomposition, where the linear layer in a transformer block is decomposed into two linear layers whose total parameter count is smaller than the original layer.
Specifically, a linear layer with a weight matrix $\bm{W} \in \mathbb{R}^{m \times n}$ is decomposed into two consecutive linear layers with weight matrices $\bm{W}_{\rm v} \in \mathbb{R}^{h \times n}$ and $\bm{W}_{\rm u} \in \mathbb{R}^{m \times h}$, respectively, where $m$, $n$ and $h$ denote the dimensions of the output, input and hidden features, respectively.
The dimension of hidden features $h$ can be dynamically tuned to allow an almost arbitrary number of parameters while maintaining structural consistency.
As $h$ is significantly smaller than the input and output dimensions $n$ and $m$, the combined parameter count of two decomposed layers is lower than that of the original layer.
Given $h$, the ratio of the number of parameters between the decomposed and original layers is $\frac{h(n + m)}{n m}$, which is numerically equal to the ratio of required multiply-accumulate operations.
In mainstream LLMs, linear layers within transformer blocks constitute the vast majority of model parameters, including the projection layers of query, key and value embeddings, and the feed-forward network (FFN).
Consequently, weight decomposition can substantially decrease the GPU memory usage and accelerate the inference of transformer-based LLMs.
Furthermore, the compression ratios for different categories of linear layers in each transformer block can be adaptively assigned to minimize performance degradation caused by linear layer decomposition.

One effective approach to initialize weights for the decomposed linear layers is singular value decomposition (SVD).
To minimize compression loss introduced by weight decomposition, we first apply data whitening to the input features \cite{SVD-LLM}.
Specifically, a forward pass is performed on a small calibration dataset to obtain the input features $\bm{X}$ of the original linear layer.
The whitening matrix $\bm{S} \in \mathbb{R}^{n \times n}$ is derived via Cholesky decomposition of the covariance matrix $\bm{X} \bm{X}^T$, such that $\bm{S} \bm{S}^T = \bm{X} \bm{X}^T$.
The whitened input features $\bm{S}^{-1} \bm{X}$ are orthonormal, thus achieving decorrelation and standardization among input features.
Subsequently, we perform reduced SVD on the enhanced weight matrix $\bm{W} \bm{S}$, expressed as:
\begin{equation}
	\bm{W} \bm{S} = \bm{U} \bm{\Sigma} \bm{V}^T ,
	\label{eq_SVD_decomp}
\end{equation}
where $\bm{U} \in \mathbb{R}^{m \times r}$, $r$ is the rank of the enhanced weight matrix, $\bm{\Sigma} \in \mathbb{R}^{r \times r}$ is a diagonal matrix containing the singular values in descending order $\left\{ \sigma_1, \sigma_2, \ldots, \sigma_r \right\}$, and $\bm{V} \in \mathbb{R}^{n \times r}$.
Given a target hidden feature dimension $h$, the top $h$ components are selected from the decomposed matrices to form $\bm{U}_d \in \mathbb{R}^{m \times h}$, $\bm{\Sigma}_d \in \mathbb{R}^{h \times h}$, and $\bm{V}_d \in \mathbb{R}^{n \times h}$.
Using these components, the initial weight matrices for the decomposed layers are given by:
\begin{equation}
	\bm{W}_{\rm u} = \bm{U}_d \bm{\Sigma}_d^{\frac{1}{2}}, \; \bm{W}_{\rm v} = \bm{\Sigma}_d^{\frac{1}{2}} \bm{V}_d^T \bm{S}^{-1}.
	\label{eq_SVD_W_u_v}
\end{equation}

We evaluate the compression loss due to SVD using the Frobenius norm of the difference between the output features of the original and decomposed layers, formulated as:
\begin{align}
	L_{\rm decomp} &= \left\| \bm{W} \bm{X} - \bm{W}_{\rm u} \bm{W}_{\rm v} \bm{X} \right\|_F^2 = \left\| \sum_{i=h+1}^r \sigma_i \boldsymbol{u}_i \boldsymbol{v}_i^T \bm{S}^{-1} \bm{X} \right\|_F^2 = \sum_{i=h+1}^r \sigma_i^2,
	\label{eq_SVD_loss}
\end{align}
where $\boldsymbol{u}_i$ and $\boldsymbol{v}_i$ denote the $i$-th column vector of $\bm{U}$ and $\bm{V}$, respectively.
As shown in (\ref{eq_SVD_loss}), the distortion of the output features caused by truncating the decomposed matrices is quantitatively determined by the squared singular values of the discarded components.
This theoretical insight enables dynamic selection of the compression ratio per layer based on the distribution of singular values, thereby improving overall model accuracy under a fixed total parameter budget.

\subsubsection{Early Exit}
\label{eesb}
The second enabling technique of familial models is early exit \cite{layerskip,shao2021branchy,rahmath2024early}, where intermediate features produced by earlier layers are directly forwarded to the language model (LM) head to predict the output token.
This design enables LLM inference on resource-constrained edge devices without significantly compromising accuracy.
The LM head consists of a layer normalization module and a linear layer that transforms the intermediate features into probabilities of all tokens in the vocabulary.
To further improve model performance at earlier exit points, we introduce lightweight trainable modules between each exit point and the LM head to refine the intermediate features before they are used for prediction, referred to as branches.
Multiple branches with varying numbers of parameters and computational complexities can be connected to a single exit point, enabling flexible adaptation to diverse hardware capabilities and task requirements.
For instance, transformer blocks whose linear layers are decomposed using different compression ratios can be inserted between the exit point and the LM head, forming a series of branches at each exit point.
Consequently, the familial model supports almost arbitrary model capabilities while maintaining consistency across different configurations.
Heterogeneous devices can adaptively choose the optimal branch based on the available hardware resources and the delay requirements of downstream tasks.

Moreover, ensuing transformer layers can continue processing the intermediate features from any exit point, opening up brand-new opportunities for cooperative inference across multiple devices.
For example, a mobile device with limited computational power may initiate inference using a lightweight branch at an early exit point.
When the network bandwidth is sufficient and the downstream task is complicated, the remaining computation is offloaded to a more powerful edge server.
The server resumes processing from the exit point to generate a response of higher quality without requiring redundant computation.
This hierarchical and collaborative inference paradigm strengthens both efficiency and accuracy in distributed computing systems.

\subsection{Implementation Strategies} \label{sec_familial_impl}

\subsubsection{Hierarchical Principal Component Decomposition}
\textbf{We propose the first implementation strategy of familial models, i.e., hierarchical principal component decomposition (HPCD), which compresses deep neural networks by decomposing weight matrices into a hierarchy of low-rank components.}
By progressively training components to fit residuals of previous stages, HPCD captures both primary and supplementary features, enabling dynamic compression.  
HPCD enables dynamic trade-offs between model accuracy and computational efficiency through progressive training and adaptive rank selection. 
Our approach achieves hardware-friendly compression while preserving critical features through residual-aware component learning.  

\textbf{Details.} We represent the weight matrix in each linear layer as a weighted sum of \( K \) low-rank components.
This decomposition can be interpreted as a matrix-valued Taylor-like series expansion, where each low-rank term progressively approximates the residual error left by the previous components. 
The HPCD training framework utilizes a hierarchical initialization strategy followed by iterative residual fitting. 
This design enables each component to sequentially capture critical features while preserving the information learned by the previous components.
For the initialization of the first low-rank component, we employ the weight decomposition method described in Section~\ref{layer_decomp}, which decomposes the linear layer weight matrices of a pre-trained model. 
For subsequent components, we adopt a sequential residual approximation approach. 
At each step $k \geq 2$, the parameters of all previously trained components are frozen, and the residual (defined as the difference between the original weight matrix and the cumulative approximation of the prior components) is computed. 
To initialize the $k$-th component, this residual matrix is approximated using the aforementioned low-rank decomposition method.
This iterative refinement process ensures that each new component enhances the overall approximation by maintaining orthogonality with respect to earlier components.
During inference, a compact weight matrix is obtained through HPCD by retaining the first-$k$ components.
The value of $k$ is adjusted based on compute or hardware constraints, allowing granular control over FLOPs and memory usage. 
We design four model variants with parameter scales ranging from 2.38B to 6.30B.
All variants share a uniform low-rank decomposition structure where each component maintains a consistent rank, while the number of components $k$ varies to achieve different parametric capacities. 
Specifically, we adapt the TeleChat \cite{he2024telechattechnicalreport} architecture by substituting the weight matrices in self-attention and feed-forward layers with HPCD-structured components.
For models with parameter sizes ranging from 2.38B to 6.30B, we select the number of components $k$ with values ranging from $1$ to $4$. 

We employ a large-scale, high-quality pre-training dataset curated through aggregation from diverse sources, establishing a comprehensive knowledge base. To preserve contextual coherence in extended sequences, we concatenate documents from the same source. Sequences are truncated and concatenated to achieve a fixed maximum context length of 8K tokens. In terms of training methodology, we utilize the Adam optimizer with $\beta_1 = 0.9$, and $\beta_2 = 0.95$. For learning rate scheduling, we implement a cosine decay schedule with a peak learning rate of $3 \times 10^{-4}$, which linearly decays to 10\% of the peak value by the end of training.  To improve training stability, we apply global gradient norm clipping of 1.0. Additionally, a weight decay of 0.01 is applied to all trainable parameters (excluding biases) to regularize the model. We employ a batch size of 4M tokens during training, with the total training token counts (in billions) ranging from $400$ to $100$.

For supervised fine-tuning data collection, we employ human annotators to create diverse and contextually rich prompts, which are then structured into coherent conversational exchanges. This process results in a dataset comprising over 100,000 fine-tuning samples, spanning multiple domains such as general Q\&A, creative writing, reading comprehension, machine translation, code generation, and mathematical reasoning. To ensure balanced domain representation, we assign resampling weights proportional to each domain’s designated importance. 
The model is initialized using the foundation model trained during the pretraining stage. Additionally, we introduce loss masks for user input questions to ensure that the loss is computed exclusively on the generated responses. 
We finetune the HPCD model family for two epochs using the SFT dataset. 
We employ a batch size of 16 samples and utilize a cosine decay learning rate schedule, initialized at $5 \times 10^{-6}$ and decaying to $1 \times 10^{-6}$. The optimizer configuration is consistent with that of the pretraining phase. During the training of the family model, we adopt a progressive strategy: earlier components are trained with their parameters frozen before moving on to subsequent components.

The HPCD training framework combines language modeling loss with knowledge distillation loss to preserve both the generative capability of the low-rank model and its alignment with the teacher model’s behavior. 
We utilize the standard autoregressive language modeling objective to minimize the cross-entropy between predicted logits and ground-truth token labels.
To accelerate the learning process, we employ a knowledge distillation framework by transferring knowledge from a pre-trained teacher model (TeleChat2-7.68B, trained on 10T tokens) to low-rank student models. This approach optimizes the alignment between the student and teacher logits distributions by minimizing the Kullback-Leibler (KL) divergence.
The total loss is a weighted sum of the two components with a tunable weighting coefficient $\lambda$ balancing both sides.
In practice, $\lambda$ is fixed at $0.4$ during training.

\textbf{Evaluations.}
We evaluate the capabilities of the HPCD model family from various perspectives using standard benchmarks. The detailed information on test benchmarks is described as follows. MMLU \cite{hendrycks2021measuringmassivemultitasklanguage} is an English benchmark comprising 57 college-level tasks. CMMLU \cite{li2024cmmlumeasuringmassivemultitask} is a Chinese benchmark evaluating large language models' knowledge and reasoning capabilities in Chinese contexts. C-Eval \cite{huang2023cevalmultilevelmultidisciplinechinese} is a comprehensive Chinese assessment containing over 10,000 questions across four difficulty tiers. GSM8K \cite{cobbe2021trainingverifierssolvemath} is a dataset of 8,500 linguistically diverse elementary mathematics problems. MATH \cite{hendrycks2021measuringmathematicalproblemsolving} is a collection of 12,500 challenging mathematics competition problems. BBH \cite{suzgun2022challengingbigbenchtaskschainofthought} is a suite of 23 challenging tasks selected from the BIG-Bench benchmark.

The detailed experimental results are presented in Table \ref{eval}. To standardize the evaluation methodology, we adopt the assessment framework provided by OpenCompass to obtain benchmark results across most datasets. Notably, despite the limited training token count, the proposed HPCD models achieve performance comparable to established models such as LLaMA2-7B, Baichuan2-7B, and ChatGLM2-6B. Moreover, the performance of each subsequent component demonstrates progressive improvement over its predecessors, indicating that later components effectively capture knowledge not acquired by earlier ones.

\begin{table}[t]
\centering
\caption{Evaluation results of the proposed HPCD familial models on standard benchmarks.}
\label{eval}
\begin{tabular}{lccccccc}
\toprule
\textbf{Model} & \textbf{MMLU} & \textbf{CMMLU} & \textbf{C-Eval} & \textbf{GSM8K} & \textbf{MATH} & \textbf{BBH}  & \textbf{Avg.}\\ 
\midrule
{LLaMA2-7B-chat} & 46.8 & 45.3 & 32.5 & 26.3 &3.3 & 38.2 & 32.1\\ 
{Baichuan2-7B-chat}& 52.8 & 54 & 55.6 &32.8&6.0& 35.8 & 39.5 \\
{ChatGLM2-6B} & 45.9 & 49.3 & 52.6 & 28.8 & 6.5&32.7 & 35.9\\
\midrule
{HPCD-rank1 (2.38B)} & 42.5 & 42.7 & 42.7 & 53.5 & 21.5&33.8 & 39.5\\
{HPCD-rank2 (3.68B)} & 45.0 & 46.3 &43.2 &56.3&26.1&40.3&42.9\\
{HPCD-rank3 (4.99B)} & 45.1 & 48.1 & 46.0 & 58.0&31.0&43.8&45.3\\
{HPCD-rank4 (6.30B)} & 47.5 & 46.5 & 47.2 & 60.8&31.0&45.7&46.5\\
\bottomrule
\end{tabular}
\end{table}

\subsubsection{Early Exiting with Scalable Branches}
\begin{figure}[!t]
    \centering
    \includegraphics[width=0.9\linewidth]{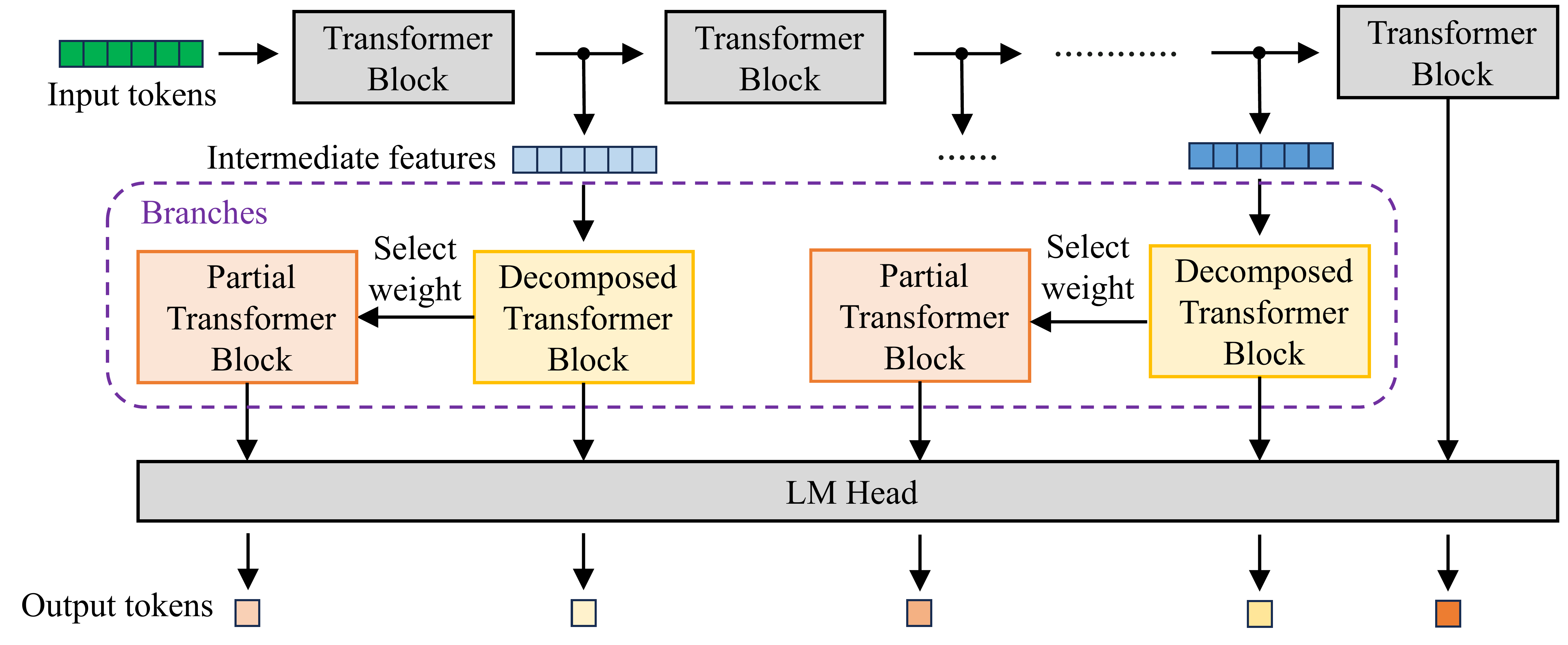}
    \caption{An illustration of the proposed EESB method.}
    \label{fig_strategy_2}
\end{figure}
\textbf{We propose the second implementation strategy of familial models, i.e.,  early exiting with scalable branches (EESB), to minimize the design cost of familial models while supporting a nearly arbitrary number of parameters.}
As illustrated in Fig. \ref{fig_strategy_2}, this strategy inserts a branch network between each exit point and the shared LM head.
The branch network is formulated as a decomposed transformer block, whose linear layers are decomposed as elaborated in Section \ref{layer_decomp}, and the total parameter count is equal to that of an original transformer block.
Based on the parameter budget, the linear layers of the transformer block in the branch network can be further decomposed with a lower compression ratio, achieved by retaining only a portion of the decomposed weight matrices.
This design enables a nearly arbitrary number of parameters while maintaining the total parameter count of a series of familial models below twice that of the main model.
{
Compared with the previous study EE-LLM \cite{chen2024eellm}, the advantage of EESB is the ability to scale the parameter budget more flexibly.
In addition to multiple exit points, each branch network contains a decomposed transformer block, where its parameters can
be dynamically tuned to achieve a more flexible tradeoff between computational cost and model performance, based on task requirements or resource constraints.
}

\textbf{Details. }
To construct a familial model with a target number of parameters, we first determine the location of the exit point as the last layer that can be executed within the parameter budget.
After the exit point, the branch is formulated by a decomposed transformer block in which all linear layers are decomposed with identical compression ratio, such that the total number of parameters of the transformer blocks prior to the exit point, the branch, and the shared LM head, is equal to the target number of parameters.
During inference, only the transformer blocks prior to the exit point and the selected branch network are loaded, substantially reducing the consumption of GPU memory and storage space to meet the hardware constraints of the deployed devices.

During training, the weight matrices of the decomposed linear layers are initialized by employing data whitening and SVD on the corresponding linear layer in the next transformer block after the exit point as in (\ref{eq_SVD_W_u_v}).
As for the shared main model, the following three training approaches are considered.
First, the main model can be frozen to minimize training costs, which is especially suitable when the downstream task is relatively simple or the available resources for training are limited.
This approach also allows for separate training of each branch, further reducing the GPU memory requirements.
Second, the main model can be fine-tuned using LoRA \cite{LoRA} to prevent overfitting when the training data is limited or biased.
Third, the entire main model can be trained end-to-end when high-quality training data and sufficient hardware resources are available.

The loss function is a weighted sum of the cross-entropy losses on both the main model and all branches.
To further optimize the performance of familial models with specific numbers of parameters, especially ones that only utilize a portion of the decomposed weight matrices, we calculate cross-entropy losses on these partial branches, which also serve as components of the loss function.
Training a large number of branches simultaneously may incur prohibitive GPU memory costs.
To address this limitation, we propose to train only a subset of branches in a single training step.
There are multiple methods to determine the selected subset, including but not limited to random selection, sequential selection, and adaptive selection based on the loss values.
Alternatively, we can train each subset of branches separately by executing only the forward pass, loss calculation, and backward pass, without performing the optimization step.
When the operations on all subsets finish, we then run the optimizer to update the parameters using the accumulated gradients.

We employ EESB to train the LLaVA-1.5-7B model~\cite{llava1.5} with 32 transformer layers and 31 possible exit points, each with a complete branch.
We construct nine partial branches exiting from the tenth to eighteenth transformer layers, each with a compression ratio of 0.75, which corresponds to total parameter counts ranging from 2.76B to 4.38B.
The main model is fine-tuned using LoRA~\cite{LoRA} with a rank of 256, the ViT and the projection layer are frozen, and all branches are trained in each optimization step by accumulating gradients from all subsets.
The model is trained for 1 epoch with the instruction tuning data used in the second training stage of the backbone model \cite{llava1.5}, consisting of 665K samples in a variety of categories, including visual question answering (VQA), optical character recognition (OCR), referring, grounding, and natural language conversation.
The hyperparameters used in training follow the settings of the second training stage of the backbone model.
We employ the AdamW optimizer with a cosine decay learning rate schedule and a peak learning rate of $2 \times 10^{-5}$.
The learning rate warmup ratio is set as 0.03, and weight decay is not applied.
The cross-entropy losses of all branches are averaged as the final loss function.



\textbf{Evaluations.}
We comprehensively evaluate the capability of the familial model by nine VQA benchmarks, including the test set of AI2D \cite{AI2D}, the English development set of MMBench \cite{MMB}, MMStar \cite{MMStar}, RealWorldQA, the test set of ScienceQA \cite{ScienceQA}, the image VQA questions in SEEDBench \cite{SEEDBench}, the test set of ChartQA \cite{chartqa}, the validation set of DocVQA \cite{docvqa}, and the validation set of InfoVQA \cite{infovqa}.
These benchmarks measure various aspects of image understanding capabilities, including perception, OCR, reasoning, mathematics, scientific knowledge, and information gathering.
We compare the proposed EESB method with the baseline model, which directly uses the LM head to generate a response from the intermediate features at each exit point.

\begin{figure}[!t]
    \centering
    \includegraphics[width=\linewidth]{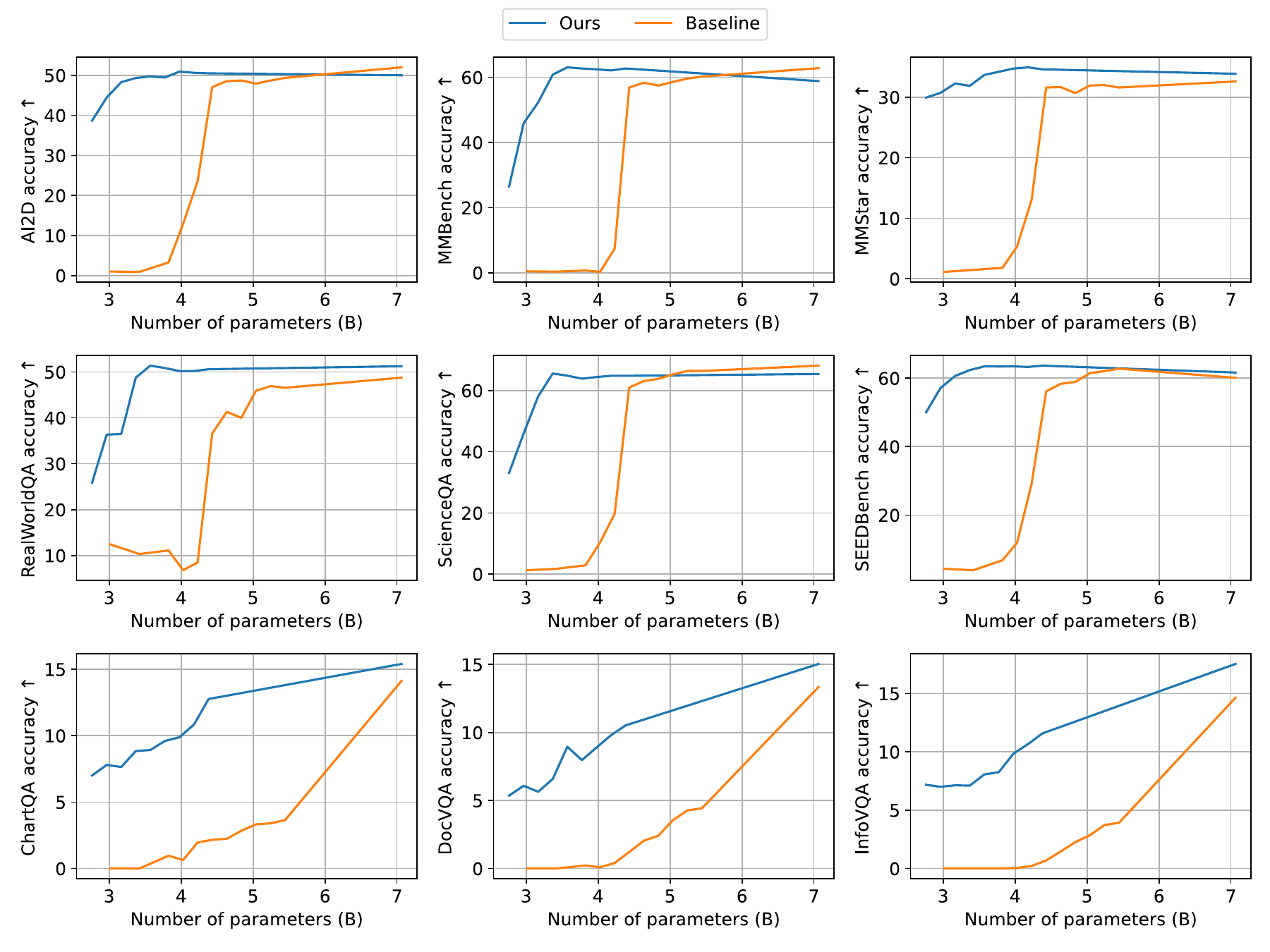}
    \caption{Evaluation results of the proposed EESB familial models on nine VQA benchmarks.}
    \label{fig_str2_vlm_tasks}
\end{figure}
Fig. \ref{fig_str2_vlm_tasks} demonstrates the accuracy on all nine benchmarks with respect to the parameter count.
For the first six general-purpose VQA benchmarks, namely AI2D, MMBench, MMStar, RealWorldQA, ScienceQA, and SEEDBench, the proposed EESB method achieves more than 95\% of the task performance with less than 45\% of the parameter count on each benchmark, compared with the 7B backbone model.
Conversely, the baseline method fails to deliver 20\% of the backbone capability until 4.03B parameters are utilized.
In terms of average performance on these six VQA benchmarks, EESB maintains 98.2\% of the backbone capability using only 3.17B parameters.
On the contrary, the baseline method activates at least 4.63B parameters to reach the target of 90\% of the backbone capability.
For the last three VQA benchmarks regarding information gathering from diagrams and tables, namely ChartQA, DocVQA, and InfoVQA, EESB achieves 82.8\% of the task performance with 4.18B parameters, while the baseline method struggles to deliver 28.6\% of the task performance with 5.44B parameters.
Although both approaches require a higher parameter count to achieve the majority of the backbone capability in information-gathering benchmarks compared to the general-purpose ones, EESB still manages to significantly outperform the baseline method.
These results underscore the effectiveness of EESB in implementing familial models, despite the significant heterogeneity of diverse VQA tasks.
On average, 
EESB significantly lowers the parameter counts required to maintain 60\% of the backbone capability from 4.43B to 2.76B.
Furthermore, EESB achieves more than 95\% of the backbone capability using only 4.38B parameters, while the baseline method only delivers 75\% of the backbone capability with 5.44B total parameters.
These results clearly show that the decomposed transformer block in the branch network further refines the intermediate features before generating a response, thereby minimizing the number of parameters necessary for delivering performance comparable to the backbone model.

\subsection{Discussions}

Guided by the valuable insights above, the familial models exhibit two advantages: highly flexible parameter configuration and enhanced efficiency in model collaboration.
\textbf{First, familial models support fine-grained adjustments on the total parameter count, attained by adaptive weight decomposition of linear layers within transformer blocks.
By carefully tuning the dimensions of hidden features between decomposed layers, a nearly arbitrary number of parameters is achievable for familial models, accommodating the hardware capabilities of heterogeneous devices.}
Given a giant model with tens of billions parameters that runs on powerful computing clusters in the cloud infrastructure, much smaller familial models can be customized to enable deployment on resource-constrained user devices and edge servers.
Apart from adapting to available computing resources, familial models are also adaptive to the diverse requirements of downstream tasks on both model intelligence and inference latency.
For instance, for delay-sensitive tasks such as simultaneous interpretation and autonomous robotic control, employing a smaller model can reduce response latency and enhance user experience, which is critical to model adoption in real-world scenarios.

\textbf{Second, familial models of varying sizes all have shared intermediate features, which can be transferred between devices and directly reused without further processing, thus enabling overhead-free split inference.}
One possible application of this feature is cooperative inference between user devices and online servers.
Specifically, the first several transformer blocks are executed on the user device, and a preliminary response is predicted based on the last activations.
These activations are transmitted to a proximate edge server for further processing to generate a more thoughtful response, which is sent back to the user.
This approach guarantees the smooth generation of model response despite potential intermittent disconnections with the server caused by a volatile wireless communication channel.
Additionally, it offloads a portion of computation from the edge server to the user device, thereby reducing the cost of AI services and increasing the concurrent requests the edge server can handle.
Another use case emerging from an aligned intermediate feature is to maximize the efficiency of speculative decoding.
The joint training of different-sized familial models renders the probability distribution of draft tokens that are generated from intermediate activations more similar to that of verified tokens predicted by the full model.
Consequently, the acceptance rate of draft tokens is higher when using a smaller familial model as the draft model, thus enhancing the inference acceleration performance.
Moreover, the shared computation between the smaller familial model and the larger one can be omitted in the inference process of the larger model.
Even when the draft model and the target model are deployed on different devices, as in the case of collaborative decoding in Section \ref{sec: hierarchical collaboration decoding}, the computation savings still persist as long as the intermediate results from the smaller model are exchanged between the two devices.

\textbf{To sum up, the intricate design of familial models has enabled efficient collaboration among diverse devices by leveraging the communication network.
Two factors contribute to this efficiency: The flexible model architecture sufficiently utilizes the available computing capacity, and the computation shared by the smaller and larger models is not repeated.}
Considering the soaring demands for model inference, the efficiency of familial models exhibits remarkable potential for widespread deployment.
Future works regarding familial models mainly entail extending the usage of familial models to more downstream tasks and deploying the trained models on a layered infrastructure.

\section{Connectivity- and Interaction-based Intelligence Emergence}\label{sec:key3}
\textbf{AI Flow facilitates the emergence of intelligence by orchestrating collaboration among heterogeneous intelligent components, thereby overcoming the limitations of isolated models through enhanced connectivity and inter-model synergy.}
By enabling hierarchical interactions among models, AI Flow integrates diverse modalities and domain-specific expertise to generate contextually coherent and globally consistent outputs, achieving synergistic capabilities that surpass the sum of individual contributions.
In this section, we explore a spectrum of collaborative paradigms among diverse model types, including LLMs, VLMs, and diffusion models, which are applied to a variety of downstream tasks.

\subsection{Collaboration among LLM Agents}\label{sec: connected agents}
\textbf{Within the AI Flow framework, LLM-based agents constitute the foundational elements whose collective intelligence emerges through progressively richer interactions. }
Existing approaches are typically within a server–server paradigm, assuming that all agents are executed on high-performance servers without the involvement of user devices.
On the contrary, AI Flow enables each user to contribute native intelligence through a personalized, lightweight LLM capable of running on mobile or personal devices.
By engaging in hierarchical interactions with models from other users or edge servers, these device-level agents can transcend their individual limitations and achieve enhanced functionality through collaboration.
Such a hierarchical architecture inherently supports scalable multi-agent collaboration, aligning well with the structure of modern distributed systems.
To allow seamless intelligence flow through the network, we propose a novel device–server collaboration paradigm that supports efficient coordination among heterogeneous LLMs and VLMs. 

\begin{figure}[t]
\centering
\includegraphics[width=0.98\linewidth]{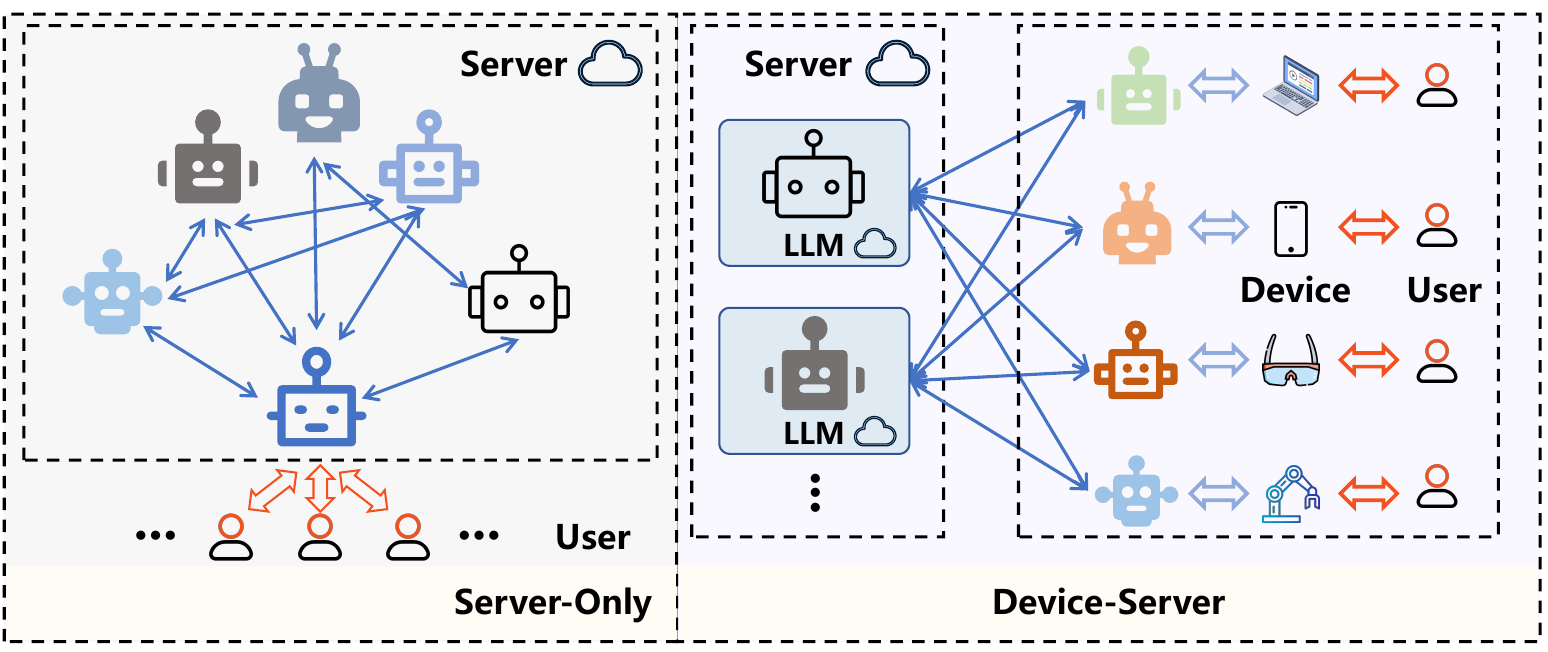}
\caption{Comparison between the device-server collaboration paradigm and the conventional server-only paradigm.}
\label{figure_CoLM}
\end{figure}

\textbf{Details.}
In the proposed scheme, compact on-device models, each optimized for a particular domain or user preference, initially preprocess the user query under constrained computational resources. 
These edge models generate preliminary responses in parallel, contributing complementary perspectives.
A central server model coordinates these user queries by the following steps:
(1) Identify the most relevant device models by matching the incoming query to descriptions on each model’s domain-specific expertise. (2) Collect individual responses from the selected models. (3) Synthesize a unified answer via a prompt-driven aggregation mechanism.
Finally, this consolidated response is broadcast back to the device models, which perform a revision step in light of the global context.
This feedback loop enhances overall coherence and accuracy by allowing edge agents to refine their outputs from their perspective while maintaining consistency with the server’s synthesis.
Besides, it allows diverse insights from all involved agents to be aggregated to achieve a consensus with enriched information.
Fig. \ref{figure_CoLM} illustrates the complete collaboration framework of our device–server architecture compared to the conventional server-only paradigm.

\textbf{Evaluations.}
We evaluate the proposed device-server architecture on both LLMs and VLMs.  
\begin{table}[t]
\centering
\captionsetup{skip=10pt}
\caption{Evaluation results of collaboration among LLMs on MT-Bench, AlpacaEval 2.0, and Arena-Hard.  
    GPT-Conv donates the abbreviation of GPT-Conversational. 
	``Agg.''   denotes the aggregated score.
    }
\label{tab:llm-results}
\renewcommand{\arraystretch}{1}
\scalebox{1.1}{
	\begin{tabular}{lccccccc}
		\toprule
		\multirow{2.5}{*}{\textbf{Model}} & \multicolumn{3}{c}{\textbf{MT-Bench}} & \multicolumn{2}{c}{\textbf{AlpacaEval 2.0}} & \multirow{2.5}{*}{\textbf{Arena-Hard}} & \multirow{2.5}{*}{\textbf{Agg.}}\\
		\cmidrule{2-4} \cmidrule{5-6}
		& 1st Turn & 2nd Turn & Avg. & LC Win & Win & & \\
		\midrule
		Qwen-Math        & 8.825  & \textbf{8.101}  & \textbf{8.465}  & 54.26 & 59.76 & 89.7 & 76.19 \\
		Qwen-Coder       & 8.650  & 7.263  & 7.956  & 45.45 & 41.91 & 78.5 & 66.66 \\
		GPT-Conv         & 5.975  & 5.709  & 5.843  & 19.71 & 17.13 & 71.2  & 48.92 \\
		DeepSeek-Creative     & 7.900  & 7.775  & 7.838  & 60.04 & 54.56 & 66.8 & 66.58 \\
		\midrule
		Ours (Qwen-Math)     & \textbf{8.900}  & 7.613  & 7.612  & 61.24 & 68.25 & \textbf{89.8} &  \textbf{78.05}\\
		Ours (Qwen-Coder)    & 8.300  & 7.557  & 7.931  & \textbf{62.72} & \textbf{69.11} & 81.3 & 76.57 \\
		Ours (GPT-Conv)      & 6.350  & 6.325  & 6.338  & 17.53 & 17.14 & 80.2  & 53.57 \\
		Ours (DeepSeek-Creative) & 8.760  & 8.089  & 8.424  & 62.70 & 71.18 &  73.2&  76.21\\
		\bottomrule
\end{tabular}}
\end{table}
In LLM experiments, we employ a set of widely used open-source language models that span a variety of representative domains. Among the simulated user agents, we utilize Qwen-Math and Qwen-Coder~\cite{qwen2025_qwen2.5} to serve as domain-specific expert models, which are specifically fine-tuned for mathematical reasoning and code generation, respectively. Moreover, two additional expert models are constructed using general-purpose models through role-based prompting: DeepSeek-Creative, with DeepSeek~\cite{deepseekai2025_deepseekv3} serving as a creative writing assistant, and GPT-Conversational, where GPT-4o~\cite{openai2024_gpt4} is configured to simulate an empathetic conversational partner. This prompt-based method allows the flexible creation of pseudo-expert agents with distinct styles or capabilities without requiring model retraining.
The intermediate outputs of these models are collectively aggregated by GPT-4o, in light of its strong reasoning capabilities and consistent performance in diverse tasks.

To evaluate the effectiveness of our approach, we select three representative benchmark datasets: MT-Bench~\cite{Bench}, AlpacaEval 2.0~\cite{AlpacaEval}, and Arena-Hard~\cite{Arena}. Table~\ref{tab:llm-results} presents the performance of collaboration among LLMs. The results demonstrate that the proposed collaboration mechanism significantly benefits weaker models. 
For example, GPT-Conversational shows a marked improvement in second-round dialogue scores on MT-Bench, indicating that context integration enhances coherence and semantic continuity in extended conversations. 
In AlpacaEval 2.0, most models achieve greater consistency with human preferences, particularly in terms of local coherence and overall win rate. 
For challenging reasoning benchmarks such as Arena-Hard, all models benefit from the collaborative setup, with performance gains especially notable in weaker models.

\begin{table}[t]
\centering
\captionsetup{skip=10pt}
\caption{
	Evaluation results of collaboration among VLMs on multiple benchmarks.
}
\label{tab:vlm-results}
\renewcommand{\arraystretch}{1.25}
\scalebox{0.95}{
	\begin{tabular}{lcccccccc}
		\toprule
		\textbf{Model} & \textbf{MME-P} & \textbf{MME-R} & \textbf{SBench} & \textbf{MMB} & \textbf{OCRBench} & \textbf{AI2D} & \textbf{MMMU-Val} & \textbf{MMMU-Dev} \\
		\midrule
		Qwen2.5-VL-7B & 1693.53 & 611.43 & 0.771 & \textbf{0.831} & \textbf{881} & \textbf{0.809} & 0.444 & 0.433 \\
		Janus-Pro-7B & 1509.38 & 270.71 & 0.701 & 0.665 & 584 & 0.679 & 0.380 & 0.373 \\
		LLaVA-1.5-7B & 1340.31 & 302.14 & 0.601 & 0.629 & 308 & 0.519 & 0.323 & 0.273 \\
		GPT-4o & 1618.96 & 672.86 & 0.755 & 0.813 & 806 & 0.737 & 0.569 & 0.567 \\
		\midrule
		Ours (Qwen2.5-VL-7B) & 1656.04 & 614.64 & \textbf{0.772} & 0.819 & 865 & 0.782 & \textbf{0.532} & \textbf{0.513} \\
		Ours (Janus-Pro-7B) & 1482.28 & 434.64 & 0.747 & 0.773 & 800 & 0.782 & 0.499 & 0.460 \\
		Ours (LLaVA-1.5-7B) & 1349.79 & 408.93 & 0.751 & 0.767 & 678 & 0.775 & 0.489 & 0.487 \\
		Ours (GPT-4o) & \textbf{1704.77} & \textbf{688.93} & 0.766 & 0.825 & 824 & 0.734 & 0.574 & 0.580 \\
		\bottomrule
\end{tabular}
}
\end{table}

In VLM experiments, we select four VLMs with distinct architectures and training paradigms as the primary evaluation targets: GPT-4o~\cite{openai2024_gpt4}, Qwen2.5-VL~\cite{Qwen2.5-VL}, Janus-Pro-7B~\cite{januspro}, and LLaVA-v1.5-7B~\cite{llava1.5}. 
To comprehensively assess their capabilities in image understanding and reasoning, we employ several representative vision-language benchmarks. 
These datasets span a wide range of skill dimensions, from basic perception to complex reasoning.
During the inference process, each model first independently generates its response. 
These outputs are then aggregated into a shared context and fed back to the original model as a prompt, guiding it to refine its answer based on the collective information. 

Table~\ref{tab:vlm-results} indicates that this mechanism leads to consistent performance gains across most tasks. 
A clear pattern emerges: the weaker the model, the more significant the improvement.
For example, LLaVA and Janus-Pro demonstrate notable improvements in text recognition and multi-step reasoning tasks. 
This suggests that when a model has perceptual or logical limitations, leveraging answers from other models can provide valuable supplementary knowledge.
For stronger models like GPT-4o and Qwen2.5-VL, performance is modestly enhanced, especially in complex reasoning tasks. 


\begin{table}[t]
\centering
\captionsetup{skip=8pt}
\caption{
    Evaluation results on the impact of increasing numbers of LLM agents on benchmark performance.
}
\label{tab:agent_numbers-results}
\renewcommand{\arraystretch}{1}
\scalebox{1.1}{
\begin{tabular}{cccc}
    \toprule
    \textbf{Number of agents} & \textbf{AlpacaEval} & \textbf{MT-Bench} & \textbf{Arena-Hard} \\
    \midrule
    1 & 4.33  & 3.77 & 3.00 \\
    2 & 7.91  & 5.26 & 7.90 \\
    3 & 14.70 & 5.33 & 10.80 \\
    4 & 14.43 & 5.45 & 12.80 \\
    5 & \textbf{15.90} & \textbf{5.50} & \textbf{14.30} \\
    \bottomrule
\end{tabular}
}
\end{table}

We also conduct experiments focusing on the number of involved LLM agents, demonstrating how collaborative intelligence emerges through their interaction.
The results on AlpacaEval, MT-Bench, and Arena-Hard are shown in Table~\ref{tab:agent_numbers-results}.
The performance improves as the number of agents increases.
For instance, the performance curve on Arena-Hard is nearly linear with respect to the number of users, indicating the practical scalability of our proposed device-server multi-agent collaboration framework.
This trend strongly supports the effectiveness of connectivity- and interaction-driven intelligence emergence within AI Flow.



\begin{table}[t]
\centering
\captionsetup{skip=10pt}
\caption{ Evaluation results on the impact of the assistant model on multimodal benchmark performance.
}
\label{tab:impact-removal_retained}
\renewcommand{\arraystretch}{1}
\scalebox{1}{
	\begin{tabular}{lcccccccc}
		\toprule
		\textbf{Model} & \textbf{MME-R} & \textbf{SEEDBench} & \textbf{MMB} & \textbf{OCRBench} & \textbf{AI2D} \\
		\midrule
            Janus-Pro-7B & 270.7 & 0.701 & 0.665 & 584 & 0.679 \\
            Ours (Janus-Pro-7B) & \textbf{434.6} & 0.747 & \textbf{0.773} & 800 & \textbf{0.782} \\
            \midrule
            Ours (w/o GPT) & 335.4 & \textbf{0.754} & 0.721 & 665 & 0.739 \\
            Ours (w/o LLaVA) & 388.9 & 0.704 & 0.629 & 782 & 0.758 \\
            Ours (w/o Qwen) & 323.9 & 0.700 & 0.606 & 684 & 0.698 \\
            \midrule
            Ours (w/ Qwen only) & 364.3 & 0.731 & 0.685 & 684 & 0.733 \\
            Ours (w/ LLaVA only) & 276.8 & 0.653 & 0.597 & 395 & 0.483 \\
            Ours (w/ GPT only) & 310.7 & 0.721 & 0.701 & \textbf{801} & 0.742 \\
		\bottomrule
\end{tabular}}
\end{table}

We also conduct ablation experiments to evaluate the role of each assistant in our collaborative reasoning framework as shown in Table~\ref{tab:impact-removal_retained}. 
The removal of any individual assistant consistently degrades performance across benchmarks, confirming that each model contributes complementary insights to the collaborative framework. Single-assistant evaluations further show that while GPT-4o performs strongest individually, no assistant alone matches the full system. These results underscore that collaboration in our design is synergistic rather than compensatory, with specialized assistants enhancing overall capabilities through collaborative reasoning.

\subsection{Collaboration among Diffusion Models}\label{sec: connected diff}
\textbf{In the AI Flow paradigm, intelligence emerges from the connectivity and interaction among distributed model components. 
By coordinating multiple models and passing information through structured connections, complex capabilities can emerge that no single model can achieve alone.}
In order to adapt to diverse network topologies and hardware capabilities, we formulate three collaboration paradigms for diffusion models: serial, parallel, and networked collaborations.
Each paradigm illustrates how hierarchical network-based collaboration and active interactions between models give rise to richer, more generalized intelligence.

\subsubsection{Serial Collaboration} 
\begin{figure*}[t]
	\centering
	\includegraphics[width=.8\linewidth]{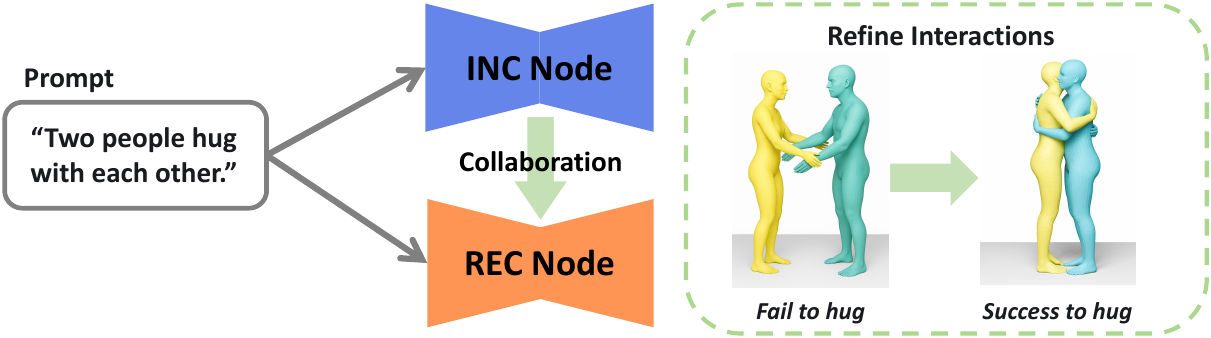}
	\caption{An illustration of the serial collaboration paradigm for diffusion models in the motion generation task. The INS node generates an initial motion sequence from a given prompt, while the REC node refines it to achieve coherent and realistic interactions.}
	\label{fig:pipeline1-1}
\end{figure*}

Benefiting from the compatibility with diverse hardware capabilities and network topologies, serial collaboration is widely adopted in real-world scenarios.
Fig. \ref{fig:pipeline1-1} illustrates a serial collaboration architecture for the motion generation task where multiple generative modules are organized in a layered pipeline to progressively construct and refine dynamic human activities. 
The system comprises two key components: the interleaved interaction synthesis (INS) module for unified motion composition and the relative coordination refinement (REC) module for enhancing cross-agent synchrony.
By working in tandem, these modules enable the emergence of complex, coordinated multi-person behaviors from modular diffusion-based generators.


\textbf{Details.}
The INS module synthesizes temporally consistent motion sequences that incorporate both solo and interactive actions. 
Rather than treating individual and social motions as disjoint modeling problems, INS learns to interleave them within a single temporal trajectory. 
This is achieved by fusing motion segments from different sources into a unified sequence, where each segment may represent either a standalone activity or a multi-person interaction. 
Temporal control signals are introduced to specify the entry points and durations of the constituting segments. 
A conditional diffusion model is then employed to generate the complete sequence, guided by text descriptions and temporal cues. The result is a motion trajectory that captures fluid transitions between self-expression and coordinated social behavior, better reflecting the way humans naturally switch between these modes.

Following the initial composition, the REC module performs interaction refinement to ensure spatial and semantic coherence among characters. 
Given a pair of motions synthesized by INS, the REC module employs a transformer-based architecture to model the spatial relationships and mutual influences between agents. 
This enforces interaction fidelity, such as ensuring that the gestures are appropriately received or mirrored by the other. 
By incorporating pairwise coordination into the generative process, the REC module corrects inconsistencies that arise from independent generation and introduces fine-grained responsiveness that reflects real-world interpersonal dynamics.
The serial design allows the system to decouple the high-level structure of motion interleaving from the low-level refinement of inter-agent dynamics. 
Each module focuses on a distinct subtask and contributes complementary knowledge to the overall synthesis process. As information flows from composition to coordination, richer interaction capabilities emerge from the combination of modular expertise. 


\textbf{Evaluations.}
We evaluate the serial collaboration on the InterHuman benchmark to assess its capacity for generating socially synchronized, text-aligned motion. As shown in Table~\ref{tab:diffusion_tab1}, our method achieves state-of-the-art performance across all key metrics. It reaches Top-1/2/3 R-Precision scores of 0.335, 0.479, and 0.584, respectively, while maintaining a low Fr\'echet inception distance (FID) of 6.332, indicating high semantic accuracy and motion realism. Compared to prior methods such as InterGen, serial collaboration improves Top-1 R-Precision by 25.3\% and reduces FID by 50.6\%, highlighting its superiority in capturing nuanced interaction dynamics.

The INS module enables precise text-to-motion alignment by jointly modeling solo and interactive components within a unified synthesis sequence. Meanwhile, the REC module refines mutual coordination between agents, reducing MM Distance by 37.0\% compared to single-person baselines like MDM. 
The serial collaboration also achieves near-human motion diversity (7.763 vs. 7.748) and optimal multimodality (1.601), confirming its ability to generate varied yet coherent motion responses.
These results underscore the benefits of structured serial collaboration toward generating physically plausible, socially aware, and semantically faithful motion sequences.

\begin{table*}[t]
	\caption{Quantitative comparison between the serial collaboration paradigm and traditional single-model paradigm on multi-human motion datasets.
        }
	\label{tab:diffusion_tab1}
	\centering
		\begin{tabular}{lccccccc}
			\toprule
			\multirow{2.5}{*}{\textbf{Method}} & \multicolumn{3}{c}{\textbf{R Precision}} & \multirow{2.5}{*}{\textbf{FID}}& \multirow{2.5}{*}{\textbf{MM Dist}} & \multirow{2.5}{*}{\textbf{Diversity}} & \multirow{2.5}{*}{ \textbf{MModality}} \\
			\cmidrule{2-4}
			& \centering \textbf{Top 1} & \centering \textbf{Top 2} & \centering \textbf{Top 3} & & & & \\
			
			\midrule
			Real  & 0.452 & 0.610 & 0.710  & 0.273 & 3.755 & 7.748 & - \\
			\midrule
			TEMOS~\cite{petrovich2022temos}  & 0.224 & 0.316 & 0.450  & 17.375 & 5.342 & 6.939 & 0.535 \\
			T2M~\cite{guo2022generating}  & 0.238 & 0.325 & 0.464  & 13.769 & 4.731 & 7.046 & 1.387 \\
			MDM~\cite{tevet2023human} & 0.153 & 0.260 & 0.339 & 9.167 & 6.125 & 7.602 & \textbf{2.355} \\
			ComMDM~\cite{shafir2023human} & 0.223 & 0.334 & 0.466 & 7.069 & 5.212 & 7.244 & 1.822 \\
			InterGen~\cite{liang2024intergen} & 0.264 & 0.392 & 0.472 & 13.404 & 3.882 & 7.770 & 1.451 \\
			FreeMotion~\cite{fan2024freemotion} & 0.326 & 0.462 & 0.544 & 6.740 & {\textbf{3.848}} & 7.828 & 1.226 \\
			\midrule
			Ours & \textbf{0.335} & \textbf{0.479} & \textbf{0.584} & {\textbf{6.332}} & 3.856 & \textbf{7.763} & {1.601}\\
			\bottomrule
		\end{tabular}
\end{table*}

\subsubsection{Parallel Collaboration}
\begin{figure*}[t]
	\centering
	\includegraphics[width=.7\linewidth]{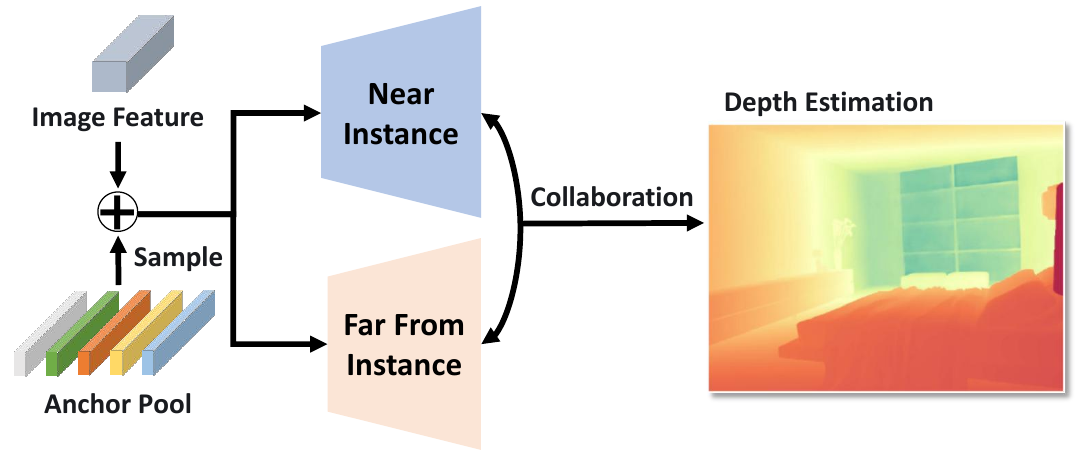}
	\caption{An overview of the parallel collaboration paradigm for diffusion models in depth estimation.
		Given an input image, a pretrained encoder extracts latent features, which are combined with a sampled anchor from an anchor pool. Two decoder branches are executed in parallel to predict scaled near-field depth, anchor validity masks, and tapered far-field depth.}
	\label{fig:pipeline3}
\end{figure*}

While the straightforward design of serial collaboration ensures low implementation costs and high compatibility, the causal dependency on prior results impedes the execution of subsequent models.
Consequently, serial collaboration is inadequate for maximizing resource utilization across a large number of interconnected devices.
To address this limitation, parallel collaboration enables simultaneous computation on multiple devices, substantially accelerating collaborative inference among models.
Fig. \ref{fig:pipeline3} introduces a framework for parallel collaboration among peer models, which leverages model diversity at the same level to tackle challenging tasks. It has the potential to address the challenge of monocular metric depth estimation across diverse environments by introducing a sliding anchor-based representation and a dual-branch architecture~\cite{wen2025_parallel}. 

\textbf{Details.}
A shared image encoder first extracts high-level features from the input image. These features are then processed by two specialized decoder branches operating in parallel: one for near-field depth and one for far-field depth.
The scaled near-depth decoder focuses on the region within a scene that lies close to the camera. By normalizing depth values with respect to a reference anchor, this branch produces detailed predictions for nearby objects. To enhance robustness, it also includes a binary mask prediction head that explicitly identifies valid depth regions within the near field. This masking mechanism enables the model to avoid unreliable predictions in ambiguous zones, leading to more stable and precise results.
The tapered far-depth decoder, in contrast, handles regions that extend beyond the anchor depth. It compresses depth values using a smooth exponential function, enabling the representation of very large distances in a normalized range. This tapering ensures that the model can represent depth values from zero to infinity, adapting to the scale variability between indoor and outdoor scenes. The outputs of both branches are reprojected into real-world depth maps and fused using the predicted mask to form a complete and continuous depth estimation.
The core component is the use of sliding anchor embeddings. During training, the model randomly samples a depth anchor from a predefined pool, allowing it to adjust its internal representation according to the scene’s depth scale. 
Smaller anchors enhance near-field resolution, while larger anchors extend the receptive field for far scenes. This dynamic adaptation enables the model to generalize effectively across varying environments without retraining.

\textbf{Evaluations.}
The parallel collaboration framework is evaluated on both NYU-V2 (indoor) and KITTI (outdoor) datasets, which are widely adopted benchmarks for metric depth estimation. 
Our model is trained jointly across datasets and fine-tuned for 70 epochs. To ensure a fair comparison with other state-of-the-art methods that impose evaluation depth caps, we follow the standard evaluation ranges: 10 meters for NYU-V2 and 80 meters for KITTI. 
Despite these constraints, our model achieves top performance across all metrics on both benchmarks, even outperforming models that are fine-tuned individually on each dataset, as demonstrated in Table \ref{tab:metric_nyu}.
This strong cross-dataset performance demonstrates the generalization ability enabled by the dual-branch design in parallel collaboration. 
The scaled near-depth branch accurately resolves local structure while the tapered far-depth branch extends predictions to distant regions. 

\begin{table}[t]
	\caption{Performance comparison between the parallel collaboration paradigm and traditional single-model paradigm on the NYU-V2 dataset.
        }
	\label{tab:metric_nyu}
	\centering
	\begin{tabular}{lcccccc}
		\toprule
		\multirow{2.5}{*}{\textbf{Method}} & \multicolumn{6}{c}{\textbf{Metric}}  \\
		\cmidrule(lr){2-7}
		~ & $\delta_1$ & $\delta_2$ & $\delta_3$ & \textbf{AbsRel} & \textbf{RMSE}  & \textbf{log10} \\
		\midrule 
		AdaBins-N~\cite{adabins} & 0.903 & 0.984 & 0.997 & 0.103 & 0.364 & 0.044 \\
		NeWCRFs-N~\cite{newcrfs}   & 0.954 & 0.981 & 0.997 & 0.113 & 0.394 & 0.083 \\
		DPT~\cite{dpt}  & 0.904 & 0.988 & 0.998 & 0.110 & 0.357 & 0.045 \\
		P3Depth~\cite{p3depth}  & 0.898 & 0.981 & 0.996 & 0.104 & 0.356 & 0.043 \\
		SwinV2~\cite{swinv2}  & 0.949 & 0.994 & 0.999 & 0.083 & 0.287 & 0.035 \\
		AiT~\cite{ait} & 0.954 & 0.994 & 0.999 & 0.076 & 0.275 & 0.033 \\
		VPD~\cite{vpd}& 0.964 & 0.995 & 0.999 & 0.069 & 0.254 & 0.030 \\
		IEBins~\cite{iebins}  & 0.936 & 0.992 & 0.998 & 0.087 & 0.314 & 0.038 \\
		ZoeDepth-N~\cite{zoedepth} & 0.955 & 0.995 & 0.999 & 0.075 & 0.269 & 0.032 \\
		DAV1-N~\cite{depthanythingv1} &  0.984 & 0.998 & \textbf{1.000} & 0.056 & 0.205 & 0.024 \\
		\midrule
		ZoeDepth-NK~\cite{zoedepth} & 0.952 & 0.995 & 0.999 & 0.077 & 0.280 & 0.033 \\
		Ours  & \textbf{0.986} & \textbf{0.998} & 0.999 & \textbf{0.049} & \textbf{0.183} & \textbf{0.021} \\
		\bottomrule
	\end{tabular}
\end{table}

\subsubsection{Networked Collaboration}
\begin{figure*}[t]
	\centering
	\begin{overpic}[width=0.6\linewidth]{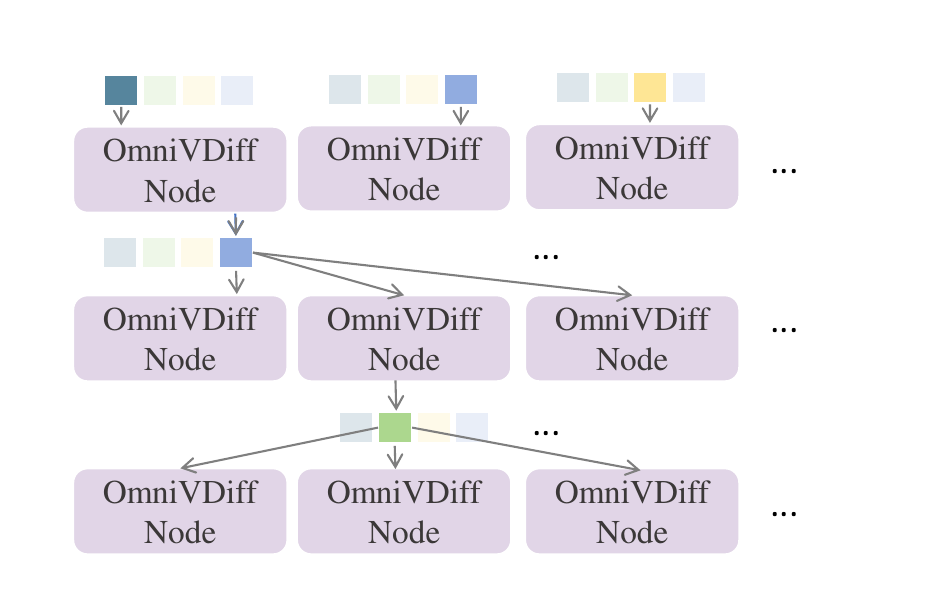}
	\end{overpic}
	\caption{An illustration of the networked collaboration paradigm.
		Different modalities are processed across multiple layers of diffusion nodes, where modal information is exchanged to support cross-modal consistency and joint reasoning.
        }
	\label{fig:pipeline2-1}
\end{figure*}

Even though parallel collaboration promotes resource utilization in the hierarchical network, it lacks flexibility in the presence of network jitter and congestion common for edge devices with wireless connections.
To achieve robustness against such fluctuations and adaptability to diverse circumstances, we introduce a networked collaboration scheme that supports arbitrary combinations of the input and output modalities.
Fig. \ref{fig:pipeline2-1} visualizes a networked collaboration scheme, namely OmniVDiff, which combines generation and understanding across multiple visual modalities within a single video diffusion framework~\cite{dianbing2025_networked}. 
Each OmniVDiff node can jointly process RGB, depth, segmentation, and Canny edge modalities. 
All modalities are concatenated along the channel dimension and perturbed with noise before being fed into the diffusion model.
The networked collaboration scheme constructs a unified generative graph by connecting these nodes.

\textbf{Details.}
OmniVDiff extends a pretrained text-to-video diffusion model by expanding the input space to accommodate multiple modalities. Each modality is first encoded into latent video patches using a shared 3D-causal variational autoencoder. These latent features, combining temporal and spatial structure across modalities, are concatenated along the channel dimension and jointly processed by a unified diffusion transformer. This architecture allows the model to learn a joint distribution over multimodal video content, capturing cross-modal dependencies and enforcing spatial-temporal alignment.
Each modality is reconstructed through its own modality-specific decoding head, ensuring that RGB, depth, segmentation, and edge modalities are faithfully recovered. These decoders are trained independently but are tightly integrated during inference through the shared diffusion backbone, allowing them to benefit from other representations. 
The output latents are decoded back into the visual domain via a pretrained 3D decoder from a variational auto-encoder (VAE) model, yielding high-fidelity video sequences.

Besides, a key feature of OmniVDiff is its adaptive modality control strategy, which dynamically determines the role of each modality based on the task. During text-to-video generation, all modalities are treated as generation targets and initialized from noise. In contrast, for X-conditioned generation tasks (e.g., depth-to-RGB), the conditioning modality is provided as a fixed input and concatenated with the noisy versions of the generation targets. 
This flexible role assignment is further guided by learnable modality embeddings, which distinguish conditioning and generation modes. These embeddings allow the model to adapt its behavior based on the task without retraining or architecture changes.
To address the scarcity of high-quality labeled video data, OmniVDiff adopts a pseudo-labeling pipeline. It generates paired data with different modalities using a series of expert models: Video Depth Anything for depth estimation, SemanticSAM and SAM2 for segmentation propagation, and the Canny algorithm for edge extraction. These labels are applied to a large-scale unannotated video dataset (Koala-36M), enabling supervised training without manual annotation. The model is trained in two stages: first, to align all modalities via text-to-video generation, and second, to perform cross-modal conditioning and understanding via the adaptive control mechanism.


\begin{table}[t]
	\caption{Performance comparison of the networked collaboration paradigm across different condition configurations.
		Evaluation metrics of various methods are Fr\'echet video distance (FVD) and kernel video distance (KVD).}
	\label{tab:comp_condgen}
	\centering
	\begin{tabular}{lccccc}
		\toprule
		\textbf{Method} & \textbf{Condition} & \textbf{FVD} & \textbf{KVD} \\
		\midrule
		T2V-zero+CtrlNet  & Text+Depth & 951.38 & 115.55 \\
		LVDM\textsubscript{ext}+Adapter & Text+Depth & 537.85 & 85.47 \\
		Make-Your-Video              & Text+Depth & 330.49 & 29.52 \\
		Ours (w/ Depth Node)             & Text+Depth & \textbf{326.99} & \textbf{27.00} \\
		Ours (w/ Canny Node)             & Text+Canny & \textbf{302.77} & \textbf{30.25} \\
		Ours (w/ Segment Node)             & Text+Segment & \textbf{254.09} & \textbf{20.05} \\
		\bottomrule
	\end{tabular}
\end{table}

\textbf{Evaluations.}
We evaluate OmniVDiff on various video generation tasks, focusing on modality-conditioned generation where the model synthesizes RGB videos conditioned on additional inputs such as depth, segmentation, or edge maps. Unlike prior works that require training separate models for each conditioning modality, OmniVDiff integrates all modalities within a single framework via adaptive modality control.
As shown in Table~\ref{tab:comp_condgen}, our model outperforms existing baselines on depth-conditioned video generation, delivering improved geometric consistency and fidelity with respect to the input depth.
These results affirm the effectiveness of the networked architecture, which unifies multiple generative and understanding tasks in a single diffusion model.
By enabling seamless transitions between generation modes and conditioning strategies, OmniVDiff exemplifies how distributed modality-specific reasoning can be orchestrated within one system.
This approach delivers scalable, flexible, and controllable video intelligence within the AI Flow framework.


\subsection{Discussions}

\textbf{AI Flow establishes connectivity and interaction as first-class primitives for the emergence of intelligence.}
Prior approaches have introduced preliminary forms of collaboration, such as prompt-based interactions, parallel cooperation among distinct model modules, and unified understanding and generation via joint model execution. 
More than this, \textbf{AI Flow generalizes these collaboration mechanisms into a hierarchical device–edge–cloud architecture, enabling balanced computational workload distribution and meeting diverse latency requirements. }
This architecture allows for seamless collaboration among heterogeneous intelligent agents, ranging from lightweight models on mobile devices to high-capacity models on powerful GPU servers.

A promising technique within this paradigm is token-level interaction, as introduced in the hierarchical collaborative decoding framework discussed in Section~\ref{sec: hierarchical collaboration decoding}. 
\textbf{This fine-grained communication and collaboration strategy allows partial inference responsibilities to be distributed across layers of the hierarchy, enabling flexible trade-offs between latency, accuracy, and computational cost.}
A representative application is the graphical user interface (GUI) agent, which requires low-latency, high-accuracy interaction with digital environments. 
In this setting, hierarchical collaborative decoding is particularly advantageous: the full decision-making process, comprising sub-tasks such as interface understanding, user intent recognition, action planning, and visual grounding, can be decomposed and distributed across the device–edge–cloud hierarchy. 
Simple and latency-sensitive sub-tasks (e.g., UI element recognition, and basic command execution) can be efficiently handled by on-device models, ensuring real-time responsiveness. 
More complex or resource-intensive sub-tasks (e.g., multi-turn reasoning, long-horizon planning) can be offloaded to edge or cloud models for enhanced accuracy and global consistency.
Through the coordinated scheduling and orchestration mechanisms provided by AI Flow, such collaborations can maintain low response latency while achieving high-quality outputs, making the system suitable for a wide range of real-world applications, including personalized assistants, real-time analytics, and multimodal interaction platforms.

Moreover, when edge devices are organized into MASs, they can collectively exhibit emergent intelligence through localized collaboration. 
These agents, each equipped with specialized LLMs or domain-specific capabilities, can coordinate their actions and reasoning to solve complex tasks that are beyond the reach of a single model. 
Under the AI Flow framework, such collaboration can be dynamically scheduled and optimized based on system context, resource availability, and task requirements.
This multi-agent paradigm is particularly well-suited for scenarios involving embodied AI, where agents embedded in physical environments (e.g., domestic robots, drones, or autonomous vehicles) must perceive, reason, and act in coordination. 
\textbf{By enabling these agents to share intermediate representations, contextual information, and feedback signals, AI Flow supports the construction of robust, adaptive, and cooperative intelligence systems capable of operating in diverse, real-world environments.}

\section{Applications}
\label{sec:app}
In this section, we explore key application scenarios of AI Flow, including embodied AI systems, wearable devices, and smart cities, which provide a comprehensive illustration of its potential to address critical challenges across specific domains. \textbf{These scenarios highlight the capability of AI Flow in enabling seamless transmission and intelligence emergence across hierarchical network architectures, driven by its core innovation of familial models as well as the foundational principle of connectivity- and interaction-based intelligence emergence.}

\subsection{Embodied AI}
Embodied AI~\cite{liu2024embodiedai, bai2024survey} represents a paradigm where intelligent systems interact with physical environments through heterogeneous agents, such as robots, autonomous vehicles, drones, and IoT devices. These agents, often characterized by diverse sensing modalities, computational capabilities, and operational constraints, collaborate to achieve complex tasks requiring perception, reasoning, and action. The essence of embodied AI lies in enabling cooperative intelligence across distributed entities, where agents must dynamically adapt to environmental changes while maintaining robust and efficient coordination. Despite its transformative potential, the challenges associated with embodied AI are manifold. First, the presence of heterogeneity among agents, ranging from hardware limitations to disparate data formats, has been identified as a key challenge. This heterogeneity has a detrimental effect on the ability to facilitate seamless communication and knowledge sharing. 
Second, the capability for real-time decision-making in dynamic environments necessitates low-latency coordination in the context of resource constraints, e.g., network bandwidth, memory storage, and processing power.
Third, it is imperative to ensure robustness against partial observability, sensor noise, and adversarial disruptions for mission-critical applications. These challenges collectively underscore the need for resilient frameworks that unify heterogeneous agents into self-organizing systems.

\textbf{AI Flow addresses these challenges by fostering collaboration among familial models, i.e., specialized AI models sharing a common architectural backbone but are tailored to the hardware constraint of each agent.} This framework facilitates efficient knowledge sharing across heterogeneous entities, thereby eliminating redundant computations. For instance, consider a scenario in which a drone and a ground robot are engaged in collaborative monitoring of a dynamic environment within a multi-agent surveillance system. By leveraging the modular architecture of familial models, both agents can fully exploit their computational capabilities through hardware-aware model adaptations. The drone is capable of capturing a wide-area view and extracting high-level motion patterns or object embeddings. These intermediate features are relayed to the ground robot for further analysis. After receiving these features, the robot can directly resume inference without repeating data preprocessing. This hierarchical exchange has been shown to minimize redundant computations, thereby reducing inference latency and computational overhead. Furthermore, by prioritizing task-relevant information sharing over exhaustive data transmission~\cite{jiawei2022_toc}, AI Flow curtails bandwidth usage and energy consumption, which are critical for resource-constrained agents operating in dynamic environments.

\textbf{Beyond the advantages offered by familial models, AI Flow leverages adaptive interaction mechanisms to enhance collective capabilities in complex tasks, leading to connectivity- and interaction-based intelligence emergence.} 
It is imperative to consider a system that integrates a robotic arm and a drone, operating in a collaborative manner for object delivery.
In this system, the robotic arm, specialized for precise object manipulation, is responsible for object identification and grasping. Concurrently, the drone, optimized for aerial navigation, positions itself to receive the object and deliver it to a remote destination. The integration of heterogeneous agents under the AI Flow framework enables tasks impossible for individual agents, showcasing the notable capability of such cohesive systems.

\subsection{Wearable Devices}
We are now witnessing a fascinating new trend in the research and development of AI-empowered wearable devices, such as smart wristbands, watches, and AI glasses \cite{zhang2024empowering}, which are capable of sensing and collecting multifaceted physiological and environmental data in a round-the-clock manner. These low-cost, handy devices targeting personal and professional use have demonstrated notable capabilities across diverse domains, including health monitoring, immersive social interaction, and environmental navigation. However, the practical challenges associated with their adoption remain significant barriers to widespread deployment. In particular, the severe constraints of on-device resources, including limited processing power, memory storage, and battery life, create a fundamental bottleneck when operating an AI model. These constraints are further compounded by the diverse functional requirements of modern applications, spanning real-time health analytics, augmented reality (AR), and multimodal environment sensing. Such applications demand high computational capacities, often creating conflicting priorities between performance, energy efficiency, and application responsiveness.

Leveraging the device-edge-cloud architecture, AI Flow emerges as a promising framework to overcome these challenges. \textbf{By dynamically partitioning computational workloads across the hierarchical network, AI Flow bridges the gap between the demanding requirements of AI-driven applications and the constrained capabilities of wearable devices.} In the context of an AR application designed for hands-free navigation, raw visual and sensor data from smart glasses can be processed locally to extract immediate spatial cues, while resource-intensive tasks such as object recognition or scene semantic segmentation are offloaded to edge servers or cloud clusters. This framework intelligently adapts to real-time constraints, prioritizing low-latency execution for interactive tasks when edge resources are available, while deferring latency-tolerant analytics to the cloud. This seamless orchestration ensures optimal utilization of heterogeneous resources, minimizing energy consumption on wearable devices while ensuring high inference accuracy and responsiveness. Hence, AI Flow empowers versatile applications to thrive within the stringent power, memory, and computational limitations inherent to wearable devices.

\subsection{Smart Cities}
Smart cities typically leverage AI-driven systems to enhance urban operations, spanning transportation, public safety, and environment monitoring. These systems depend on interconnections between heterogeneous entities, including IoT sensors, autonomous vehicles, and aerial drones. A pivotal innovation in modern urban development is the low-altitude economy, where drones and urban aerial mobility systems are synergistically integrated within regulated low-altitude airspace to execute critical tasks such as emergency medical deliveries, infrastructure inspections, and traffic management. At the heart of this paradigm is the seamless integration of AI into dynamic urban ecosystems, enabling real-time, data-driven decision-making that harmonizes urban efficiency with environmental sustainability. 
However, the challenges of implementing a smart city are multidimensional. Real-time decision-making necessitates ultra-low latency coordination to address dynamic scenarios, such as rerouting autonomous vehicles during emergencies or dispatching drones for urgent medical deliveries, all under stringent energy and bandwidth constraints. Furthermore, urban systems must scale to accommodate the exponential proliferation of connected devices and the operational complexity of the low-altitude economy, where hundreds of drones may coexist with traditional infrastructure in shared airspace.

\textbf{To address these challenges, the hierarchical network coordination framework of AI Flow enables the dissemination of intelligence across end devices, edge servers, and cloud clusters.}
Familial models, optimized for specific tiers within this hierarchy, facilitate efficient data aggregation and distributed decision-making. At the device tier, resource-constrained devices (e.g., traffic cameras and delivery drones) execute lightweight models tailored for localized tasks. For instance, drones in low-altitude delivery networks utilize edge-optimized models for obstacle navigation and collision avoidance, transmitting only summarized trajectory data to edge servers. At the edge tier, localized AI models aggregate multi-modal data (e.g., traffic flow from drones, vehicles, and pedestrian sensors) to optimize regional decisions, such as dynamic traffic signal adjustments or drone rerouting during peak hours. The cloud tier integrates city-wide data to train global models, predict systemic trends (e.g., congestion patterns), and disseminate model updates to edge servers, ensuring continuous system-wide adaptation.

\textbf{Through the framework of AI Flow, smart cities achieve a holistic integration of AI-driven intelligence, enabling seamless coordination across heterogeneous systems while addressing critical urban challenges.} The hierarchical architecture of AI Flow ensures prompt decision-making by distributing computational tasks across the hierarchical network architecture, minimizing bottlenecks, and enhancing responsiveness in dynamic scenarios.

\section{Future Work}
\label{sec:future}
Building on the demonstrated effectiveness of AI Flow across critical applications, this section outlines its prospective research directions. \textbf{Key focus areas include federated learning, distributed edge inference, and adaptive network orchestration, aiming to advance the capabilities of AI Flow in real-world deployment.}


\subsection{Federated Learning}

The development of intelligent AI systems relies heavily on extensive, high-quality training data. 
\textbf{However, industry predictions warn that future advancements could be impeded by a growing scarcity of such data, as current consumption rates increasingly surpass new data generation rates.
To sustain progress, leveraging private data scattered across users is critical for AI advancement.}
This necessitates distributed training approaches to update models without compromising data privacy.
Federated learning (FL) \cite{shao2022dres,yang2019federated,huang2025keeping,jiawei_fl} addresses this by enabling collaborative model optimization across decentralized devices while preserving privacy. Instead of sharing raw data, FL systems exchange model updates (e.g., gradients or parameters), which are aggregated to refine a global model.
However, in the era of large models, FL encounters challenges in computational, memory, and communication overhead. Specifically, pretraining or full-parameter fine-tuning of LLMs is computationally demanding, requiring substantial memory allocation and processing power, yet most FL participants operate on consumer-grade hardware. Moreover, the frequent exchange of high-dimensional model parameters across distributed devices and the server introduces prohibitive communication costs. Given that FL typically demands multiple rounds of synchronization, this cumulative overhead severely limits its scalability and efficiency. 

To tackle these challenges, parameter-efficient fine-tuning (PEFT) methods have emerged as a promising solution, significantly reducing hardware demands and alleviating communication burdens. Although these approaches enhance communication efficiency, they inevitably sacrifice model capability when introducing update bias during training. For instance, techniques like LoRA constrain updates to compact parameter subsets, distorting gradient information and degrading the performance of models. 
Therefore, developing a holistic FL framework that balances parameter efficiency with model capability remains a critical research goal.

\subsection{Distributed Edge Inference}
Distributed edge inference \cite{letaief2021edge} involves partitioning large AI models into smaller modules, enabling their collaborative execution across interconnected edge devices. 
\textbf{This paradigm leverages decentralized computational resources to achieve scalable and low-latency deployment, strategically mitigating individual device limitations like constrained memory and processing power through networked coordination.} However, implementing these systems in real-world edge environments faces formidable obstacles. First, edge devices exhibit diverse computational capabilities, memory constraints, and energy limitations. Efficiently partitioning models across heterogeneous devices while balancing workload reasonably remains a critical hurdle.
Second, edge environments are inherently unstable as devices frequently join or disconnect from the network, resulting in highly volatile workloads.
Third, distributed edge inference often incurs significant inter-device data transfer costs, e.g., high-dimensional multi-modal source data and intermediate features, which dominate end-to-end latency and bottleneck system scalability.

Although current data and model parallelism schemes have been widely adopted for collaborative inference, they often assume homogeneous infrastructure and stable network conditions, yet such an assumption rarely holds in edge environments. Hence, traditional parallelism struggles with device heterogeneity and intermittent disconnections, leading to bottlenecks in synchronization and load balancing. Future work will focus on developing efficient distributed edge inference systems capable of dynamic reconfiguration in response to heterogeneous environments.

\subsection{Adaptive Network Orchestration}
\textbf{The emergence of intelligence in distributed systems increasingly depends on collaborative learning across hierarchical network architectures.} However, as the scale of deployment expands, network conditions among participants become highly heterogeneous, ranging from high-speed communication across cloud servers to low-bandwidth mobile connections at the network edge. This variability complicates synchronous collaboration and necessitates frameworks that can adapt to differing communication capabilities. 
Ubiquitous intelligence faces critical challenges due to network heterogeneity. In particular, participants in the distributed system operate under varying network conditions, leading to unpredictable latency and bandwidth fluctuations. The resulting topology complexity introduces a degree of difficulty in terms of the robust coordination required across the network. Besides, the dynamic network conditions exacerbate the orchestration complication. In mobile communication scenarios, participants frequently encounter intermittent disconnections due to device mobility or changing wireless conditions. Besides, edge-cloud communication links are prone to network jitter and variable delays, which can degrade the reliability of data exchange and model synchronization. Addressing these issues demands adaptive communication protocols and resilient system designs that maintain performance despite time-varying network states.

Wireless ad hoc networks (WANETs) offer a promising decentralized architecture to address the dynamic conditions of edge networks. However, they are not sufficient to meet the requirements of ubiquitous intelligence. This is due to the fact that the ubiquitous intelligence systems necessitate seamless coordination across the hierarchical network architecture. Thus, the development of a network-aware intelligence framework to bridge the hierarchical resources and adapt to the diverse network conditions is a critical research goal.

\section{Conclusions}
\label{sec:conclusion}
In this paper, we present AI Flow, a novel framework integrating cutting-edge AI and communication network technologies to address the critical challenges of hardware resource limitations and communication constraints in deploying large AI models at the network edge. 
AI Flow emphasizes three key points as its core: device-edge-cloud collaboration, familial models, and connectivity- and interaction-based intelligence emergence.
In particular, AI Flow establishes a paradigm shift for ubiquitous AI by synergizing device-edge-cloud architecture with familial model collaboration, ensuring timely responsiveness and ubiquitous accessibility for intelligent systems on resource-constrained edge devices. 
Through innovative connection and interaction paradigms among agent models (e.g., LLMs, VLMs, diffusion), AI Flow attains emergent intelligence surpassing the capacity of any single model and enables powerful AI applications at the network edge. Moreover, we provide a comprehensive analysis of key application scenarios of AI Flow, revealing its capabilities in real-world edge AI solutions. Finally, future work directions are outlined to expand the scope and capabilities of AI Flow.
This paper aims to present the conceptual foundation, core mechanisms, and transformative potential of AI Flow, with a focus on the integration of AI, communication, and network to demonstrate its capacity for enabling ubiquitous AI capabilities at the network edge.

\ifCLASSOPTIONcaptionsoff
  \newpage
\fi

\section*{Abbreviations}
\begin{table*}
\centering
    \begin{tabular}{cc}
    \toprule
        Abbreviation & Definition \\ \hline
        AI & Artificial Intelligence \\
        IT & Information Technology \\
        CT & Communication Technology \\
        LLM & Large Language Model \\
        IoT & Internet of Thing \\
        VLM & Vision-Language Model \\
        ML & Machine Learning \\
        SVM & Support Vector Machine \\
        RL & Reinforcement Learning \\
        RNN & Recurrent Neural Network \\
        CNN & Convolutional Neural Network \\
        CoT & Chain-of-Thought \\
        MAS & Multi-Agent System \\
        GB & Gigabyte \\
        BS & Base Station \\
        RSU & Roadside Unit \\
        TOFC & Task-Oriented Feature Compression \\
        DPC-KNN & Density Peaks Clustering Based on $K$ Nearest Neighbors \\
        PDF & Probability Density Function \\
        STE & Straight-Through Estimator \\
        MLP & Multi-Layer Perceptron \\
        FFN & Feed-Forward Network \\
        SVD & Singular Value Decomposition \\
        LM & Language Model \\
        HPCD & Hierarchical Principal Component Decomposition \\
        KL & Kullback-Leibler \\
        EESB & Early Exiting with Scalable Branches \\
        VQA & Visual Question Answering \\
        OCR & Optical Character Recognition \\
        INS & Interleaved Interaction Synthesis \\
        REC & Relative Coordination Refinement \\
        FID & Fréchet Inception Distance \\
        FVD & Fréchet Video Distance \\
        VAE & Variational Autoencoder \\
        KVD & Kernel Video Distance \\
        GUI & Graphical User Interface \\
        AR & Augmented Reality \\
        FL & Federated Learning \\
        PEFT & Parameter-Efficient Fine-Tuning \\
        WANET & Wireless Ad Hoc Networks \\
    \bottomrule
    \end{tabular}
    \caption{List of abbreviations and their definitions.}
    \label{tab:abbreviations}
\end{table*}
Table \ref{tab:abbreviations} is a list of abbreviations used throughout this paper.

\section*{Declarations}
\subsection*{Availability of data and material}
While most of the datasets used for training are publicly available, a portion of training data for HPCD models originates from proprietary datasets and cannot be shared due to legal and contractual restrictions. These restricted datasets were used exclusively for model training and do not affect the reproducibility of the core findings presented in this paper. Source data are provided with this paper.

\subsection*{Code availability}
Code for the implementation of familial models is publicly available on GitHub at \url{https://github.com/TeleAI-AI-Flow/AI-Flow-Ruyi}. 

\subsection*{Competing interests}
The authors declare no competing interests.

\subsection*{Funding}
The authors declare no applicable fundings.

\subsection*{Authors' contributions}
Xuelong Li initiated and led this research project. 
Jiawei Shao facilitated this work. 
Hongjun An, Wenhan Hu, Sida Huang, Siqi Huang, Ruanjun Li, Yiliang Song, Zihan Wang, and Cheng Yuan contributed to the experiments with the assistance from Jiawei Shao, Hongyuan Zhang, and Yuanzhi Liang, under the guidance of Chi Zhang. 
Wenhao Zhuang, Ruanjun Li, Yuanzhi Liang, Cheng Yuan, Siqi Huang, Zihan Wang, and Jiawei Shao wrote this paper.

\subsection*{Acknowledgements}
We thank Yunuo Hu for assistance in the illustration of some figures, and Chen Chen for assistance in the revision of some references.

\bibliographystyle{IEEEtran}
\bibliography{ref}

\end{document}